%% file: hicss_GPR_arxive.tex
\documentclass[10pt]{article}
\input{hicss51-packages.tex}
\usepackage[fleqn]{amsmath}
\usepackage{mathptmx}
\usepackage{subfig}

\title{Electric Load and Power Forecasting Using Ensemble Gaussian Process Regression}

\author{Tong Ma \\
  Pacific Northwest \\National Laboratory\\
   \\\And
  Renke Huang\\
  Pacific Northwest \\National Laboratory\\
  \\\And
  David Barajas-Solano\\
  Pacific Northwest \\National Laboratory\\
  \\\And
  Ramakrishna Tipireddy\\
 Pacific Northwest \\National Laboratory\\
  \\\And
 Alexandre Tartakovsky\\
 Pacific Northwest \\National Laboratory\\
  {\underline{Alexandre.Tartakovsky@pnnl.gov}}}
 
\date{}

\begin{document}
\maketitle
\begin{abstract}
We propose a new forecasting method for predicting load demand and generation scheduling.  Accurate week-long forecasting of load demand and optimal power generation is critical for efficient operation of power grid systems. In this work, we use a synthetic data set describing a power grid with 700 buses and 134 generators over a 365-days period with data synthetically generated at an hourly rate.  The proposed approach for week-long forecasting is based on the Gaussian process regression (GPR) method, with prior covariance matrices of the quantities of interest (QoI) computed from ensembles formed by up to twenty preceding weeks of QoI observations. Then, we use these covariances within the GPR framework to forecast the QoIs for the following week. We demonstrate that the the proposed ensemble GPR (EGPR) method is capable of accurately forecasting weekly total load demand and power generation profiles.
The EGPR method is shown to outperform traditional forecasting methods including the standard GPR and autoregressive integrated moving average (ARIMA) methods.
\end{abstract}

\section{Introduction}
\label{sec:intro}

Accurate short-term forecasting of electric power generation and load demand plays an essential role in the control and planning of electric power grid systems \cite{t1}. For example, an overestimation of power generation and load demand leads to excess energy consumption, while underestimation may result in blackouts. Forecasting of dynamic processes, including electric power and loads, is difficult because such systems are directly or indirectly affected by a large number of uncertain and uncontrollable factors. 

Existing forecasting techniques are typically classified into two categories: statistical techniques and artificial intelligence (AI) techniques. Statistical techniques include multiple linear regression (MLR) models~\cite{t2,t3,t4,t5}, semi-parametric additive models~\cite{t6,t7,t8,t9}, autoregressive integrated moving average (ARIMA) models~\cite{t10,t12}, and exponential smoothing models~\cite{t10,t11}.
AI techniques include artificial neural network (ANN)~\cite{t13}, fuzzy regression models~\cite{t14,t15}, support vector machines (SVMs)~\cite{t16}, gradient boosting machines~\cite{t17,t18}, and Gaussian process regression (GPR)~\cite{t19,t20,t21}.
In the standard GPR method, a time series is assumed to be a realization of a Gaussian process with prescribed parameterized mean and covariance functions.
The parameters of these mean and covariance functions are learned from the timeseries measurements by maximizing the marginal likelihood function of the measurements or other pseudolikelihoods.

Here, we present a new forecasting method, which we call the ensemble GPR (EGPR), and apply it for weekly forecasting of load demand and generation scheduling of a power grid with 700 buses and 134 generators. 
In EGPR, we compute the covariance functions of the quantities of interest (QoI: here, load demand and generation scheduling) from the timeseries of QoI measurements, which we treat as ergodic~\cite{feller2008introduction}, i.e., we treat the $N$ weeks prior to the beginning of the forecast as realizations of the same Gaussian process.
Then, the mean and covariance are computed as the ensemble statistics of ensemble formed by these $N$ weeks.
$N$ cannot be too large as the statistics could be affected by seasonal variations, thus violating the assumption of ergodicity.
On the other hand, $N$ cannot be too small as to result in inaccurate, noisy ensemble statistics.
In the considered data set we find that Mondays are poorly correlated to the rest of the week and $N=20$ is required for accurate weekly forecast using data collected on Mondays.
We also find that the remaining days of the week are more strongly correlated with each other.
Therefore, accurate forecast can be obtained using data collected on Tuesdays by choosing $N=10$.
We demonstrate that the the proposed ensemble EGPR outperforms traditional forecasting methods, including the standard GPR and ARIMA methods.

The proposed EGPR method is conceptually similar to the physics-informed GPR method~\cite{tartakovsky2019physicsHICSS,YangJCP,yang2018physics}.
In physics-informed GPR, ensemble realizations of the QoIs are generated by repeatedly sampling stochastic models of the dynamics to be forecasted.
For example, in \cite{tartakovsky2019physicsHICSS} physics-informed GPR was used for the forecasting of the dynamics of a single-machine-infinite-bus (SMIB) system powered by random mechanical wind power fluctuations.
The mechanical wind power was treated as a random process and the ensemble was generated by repeatedly solving the SMIB swing equations.
An open question is how to apply this ensemble-based approach to GPR for forecasting problems where either there is no stochastic model or the stochastic model is computationaly costly to evaluate, but historical timeseries observations of QoIs are available.
One such problem is the forecasting of load demand and generation scheduling for large power grid systems, such as the 700-buses, 134-generators considered in this work.
For this problem, repeatedly solving the power flow and economic dispatch problems is computationally costly, and we aim to calculate forecasts using historical data.      

This paper is organized as following. Section 2 describes the synthetic dataset of load demand and generation scheduling for a 700-bus, 134-generator system.
The EGPR method is presented in Section 3.
Section 4 presents the application of EGPR to weekly forecasting of load demand and generation scheduling.
Comparisons between EGPR, standard data-driven GPR and ARIMA are also presented for weekly total load forecasting.
Finally, conclusions are presented in Section 5.

\section{Synthetic data set}
\label{sec:synthetic}

In this work, we use a synthetic dataset describing a power grid with
700 buses and 134 generators over a 365-days period, with measurements of load and power generation for each generator reported every hour.  

The synthetic dataset is constructed as follows:
In the first step, we generate the chronological system-level total load for one year with one-hour time resolution.
Then, the ``base case'' power flow is generated as described in~\cite{k1}.
The historical Duke Energy hourly load shape~\cite{k2} is used to develop hourly load values with the same participation factor as the original ``base case'' power flow.
The peak load of the system is 12,926 MW, occurring at Dec. 17, 18:00.
The generator parameters for the unit commitment (UC) include the generator fuel cost curves, generator minimum and maximum real power output, and generator minimum and maximum ramping up and down rate.

In the second step, the total system load profile and the generator parameters are fed into an UC and hourly dispatch program, which outputs the on/off status and the real power output of each generator with one-hour time resolution.
In the third step, the chronological AC power flow is computed for the power grid using the PSS/E software to generate different power flow scenarios with one-hour time resolution. The chronological system-level load profile is disaggregated to produce load for each bus in the system and is used as input to PSS/E, together with the on/off status and real power of each generator from the UC and hourly dispatch results.
PSS/E provides a converged power flow solution for each scenario.
In the final forth step, we reinforce scenarios by adjusting voltage and line flows, multi-block switchable shunts, and lines.

We use this data set to demonstrate and validate the proposed EGPR methods for weekly forecast of the total load and generation scheduling (specifically the real power output of each generator; the on/off status of each generator is not considered in this work). 

\section{Ensemble GPR}

In this section we described the proposed EGPR method for forecasting states of the power grid using historical observations of the states.
To introduce the EGPR method, we define the vector $x = [x^{o},x^{f}]^T$, where  $x^{o} = [x^{o}_1,...,x^o_{N^o}]^T$ ($x^o_i= x^o(t_i)$) is the vector of observed state measurements, and $x^{f} = [x^{f}_1,...,x^f_{N^f}]^T$ ($x^f_i= x^f(t_i)$) is the vector of predicted state values. 
The GPR method assumes that $x^0$ and $x^f$ are realizations of the Gaussian process  $X = [X^{o},X^{f}]^T$
\begin{equation}
  \label{e1}
  \begin{bmatrix}
    X^o \\ X^{f}
  \end{bmatrix} \sim \mathcal{N} \left (
    \begin{bmatrix}
      \bar{X}^o \\ \bar{X}^{f}
    \end{bmatrix},
    \begin{bmatrix}
      K_{oo} & K_{of} \\
      K_{of}^T &  K_{ff}
    \end{bmatrix}
  \right ),
\end{equation}
where   $\bar{X}^o$ and $\bar{X}^{f}$ are the so-called prior (or, unconditional) mean of $X^0$ and $X^f$, respectively, and $K_{oo}$, $K_{of}$, $K_{of}^T$, and $K_{ff}$ are the  prior (cross)covariances between  $X^o$ and $X^o$,  $X^o$ and $X^f$,  $X^f$ and $X^o$, and  $X^f$ and $X^f$, respectively.

Given $x^{o}$, $x^f$ is forecasted as
\begin{equation} \label{e2}
	x^{f} = \bar{X}^f+K_{of}^T K_{oo}^{-1}\left(x^o-\bar{X}^o\right)
\end{equation}
with forecast (or \emph{posterior}) covariance given by
\begin{equation} \label{e3}
	k_{ff} = K_{ff} - K_{of}^T K_{oo}^{-1}K_{of}.
\end{equation}
The diagonal elements of the covariance matrix $k_{ff}$ are equal to the posterior variance of the forecasted state $x^{f}$ at different times and describe the uncertainty in the forecast. 

Estimation of the prior statistics ($\bar{X}^o$, $\bar{X}^{f}$,$K_{oo}$, $K_{of}$, $K_{of}^T$, and $K_{ff}$) is the main challenge in GPR.
The standard GPR method assumes parameterized forms for the prior statistics, and the parameters of these parameterized forms are estimated by maximizing a the marginal likelihood or another pseudolikelihood of the measurements $x^o$ using \cite{t19,t20,t21}.
In Section \ref{ss4}, we use the common ``Gaussian'' or ``Squared Exponential'' parameterized form of the covariance function and demonstrate that under such assumption the GPR method fails to accurately predict the total load of the considered power grid.

In the EGPR, we assume that the power grid data is statistically ergodic, e.g., for a weekly forecast it is possible to divide the historical data into subsets of weekly data and treat each subset as well as the forecasted week as  realizations of the same random process. Then, the prior statistics can be computed from these subsets as ensemble statistics.   
To illustrate the proposed EGPR method, we consider a problem of forecasting the total load for Monday-Sunday of the $Y$th week using hourly measurements of the load for the preceding $N$ weeks. Let $l^o = [l^o_1, l^o_2,..., l^o_{24}]^T$ denote the vector of observed hourly loads on Monday of the $Y$th week and let $l^f = [l^f_{25}, l^f_{26},...,l^f_{168}]^T$ (where 168 is the number of hours in a week)
denote the vector of hourly forecasted load for Tuesday-Sunday of the $Y$th week.
To estimate the prior statistics, we form the ensemble from the previous $N$ weeks as $L^o_i = [L^o_{i,1}, L^o_{i,2},..., L^o_{i,24}]^T$ and $L^f_i = [L^f_{i,25}, L^f_{i,26},...,L^f_{i,168}]^T$ ($i=1, 2,...,N$), where $L^o_{i,k}$ is the measured load at the $k$th hour (on Monday) of the $i$th week ($k=1,24$ and $i=1,N$), and  $L^f_{i,k}$ is the measured load at the $k$th hour of the $i$th week ($k=25,168$ and $i=1,N$).   
Then, the unconditional statistics can be computed as 
\begin{equation}
	\bar{L}^o_k = \frac{1}{N} \sum_{i=1}^N L^o_{i,k}, \quad k = 1, 2,..., 24,
\end{equation}
\begin{equation}
	\bar{L}^f_k = \frac{1}{N} \sum_{i=1}^N L^f_{i,k}, \quad k = 25, 26,..., 168,
\end{equation}
\begin{equation}
\begin{aligned}
K_{oo,kl}& =\frac{1}{N-1} \sum_{i=1}^N (L^o_{i,k}-\bar{L}^o_k)(L^o_{i,l}-\bar{L}^o_{l})\\
&k=1,...,24,\quad l=1,...,24,
\end{aligned}
\end{equation}
\begin{equation}
\begin{aligned}
K_{of,kl}& =\frac{1}{N-1} \sum_{i=1}^N (L^o_{i,k}-\bar{L}^o_k)(L^f_{i,l}-\bar{L}^f_{l})\\
&k=1,...,24,\quad l=25,...,168,
\end{aligned}
\end{equation}
and
\begin{equation}
\begin{aligned}
K_{ff,kl}& =\frac{1}{N-1} \sum_{i=1}^N (L^f_{i,k}-\bar{L}^f_k)(L^f_{i,l}-\bar{L}^f_{l})\\
&k=25,...,168,\quad l=25,...,168.
\end{aligned}
\end{equation}

The forecast of the load $l^f$ and its posterior covariance can be found from using Eqs~(\ref{e2}) and~(\ref{e3}) and by setting $x^o=l^o$ and $x^f=l^f$.
As discussed in Section~\ref{sec:intro}, the choice of $N$ is critical to the performance of EGPR.
Specifically, $N$ cannot be too large enough to violate the assumption of ergodicity, and $N$ cannot be too small as to result in inaccurate prior statistics.
In this work we select $N$ by analyzing the structure of the ensemble covariance and the difference between the forecast and the prior mean, as described in Section~\ref{sec:results}.

\begin{figure}[tbh]
  \centering%
  \includegraphics[width=\linewidth]{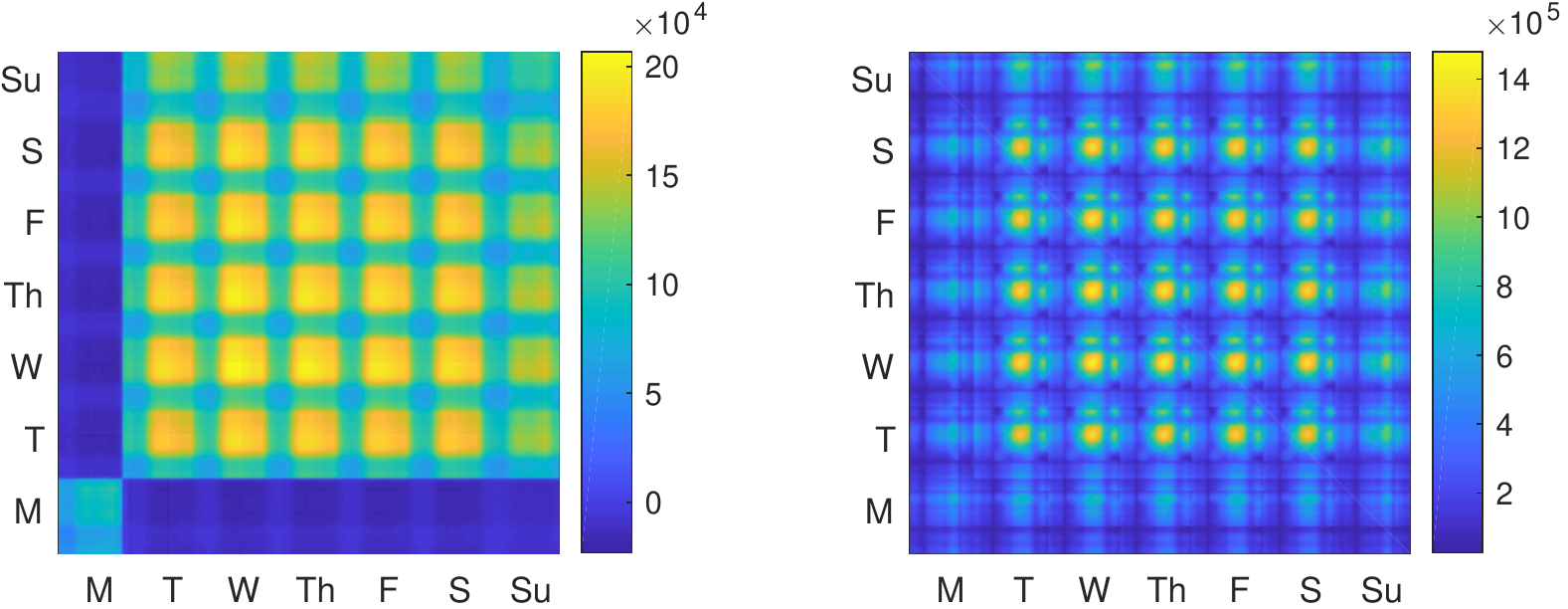}
  \caption{Empirical covariance for the week of 6/9--6/15 computed using $5$ realizations (left) and $20$ realizations (right)}
  \label{fig:load-forecast-covar}
\end{figure}

\section{Weekly forecasting of load demand and generation scheduling}
\label{sec:results}

In this section, we present results for weekly forecasting of total load demand and generation scheduling. Specifically, we forecast the total load demand and generation scheduling for Generator 15 for the weeks of 6/9--6/15, 8/11---8/17, 10/13--10/19, and 12/15--12/21. These four weeks represent the seasons of summer, fall, and early winter.
For the case of total load demand, we present a comparison between the forecasting performance of EGPR, standard GPR, and ARIMA.

\begin{figure*}[tbh]
  \centering%
  \subfloat[6/9--6/15]{\includegraphics[width=0.45\textwidth]{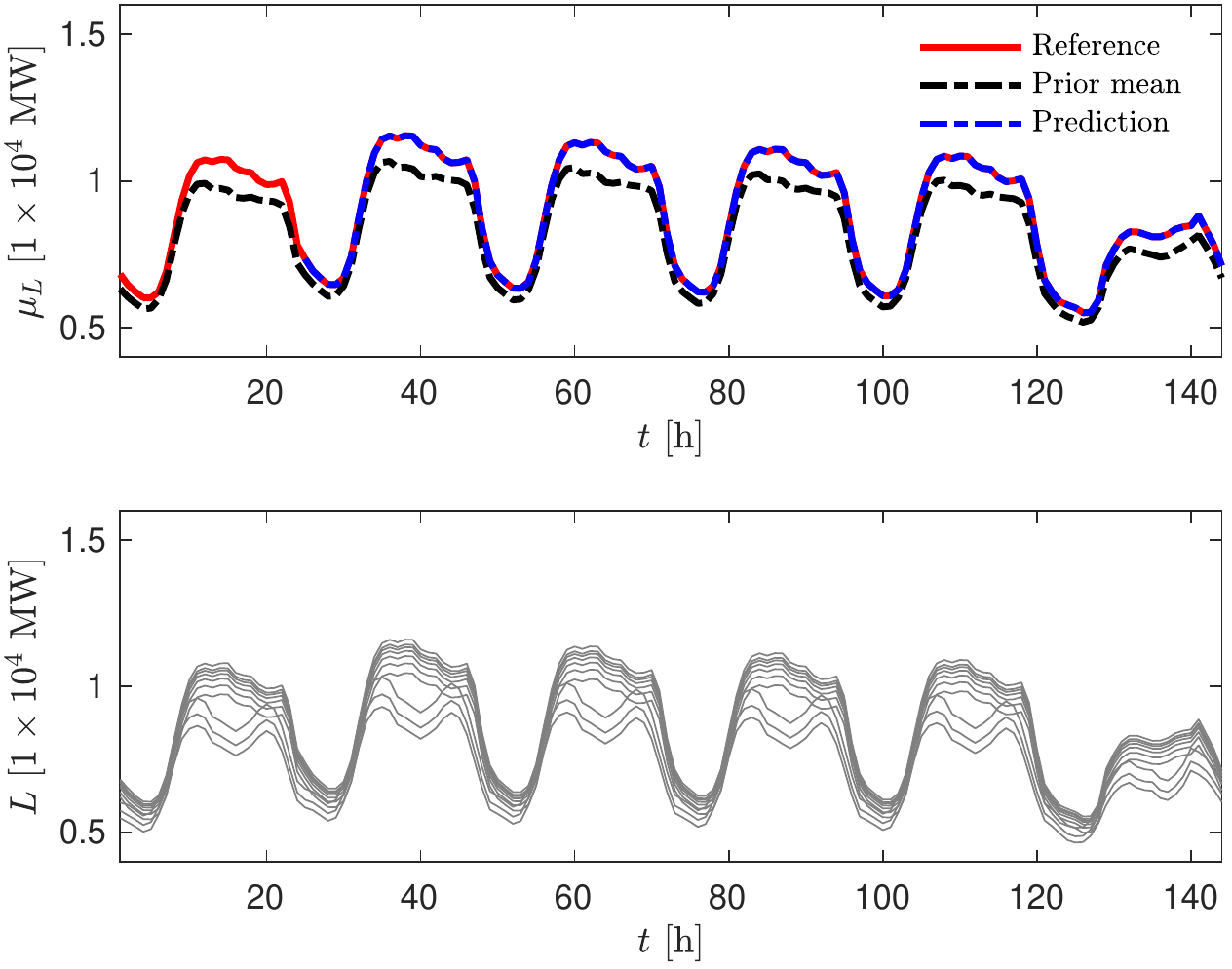}}\ %
  \subfloat[8/11--8/17]{\includegraphics[width=0.45\textwidth]{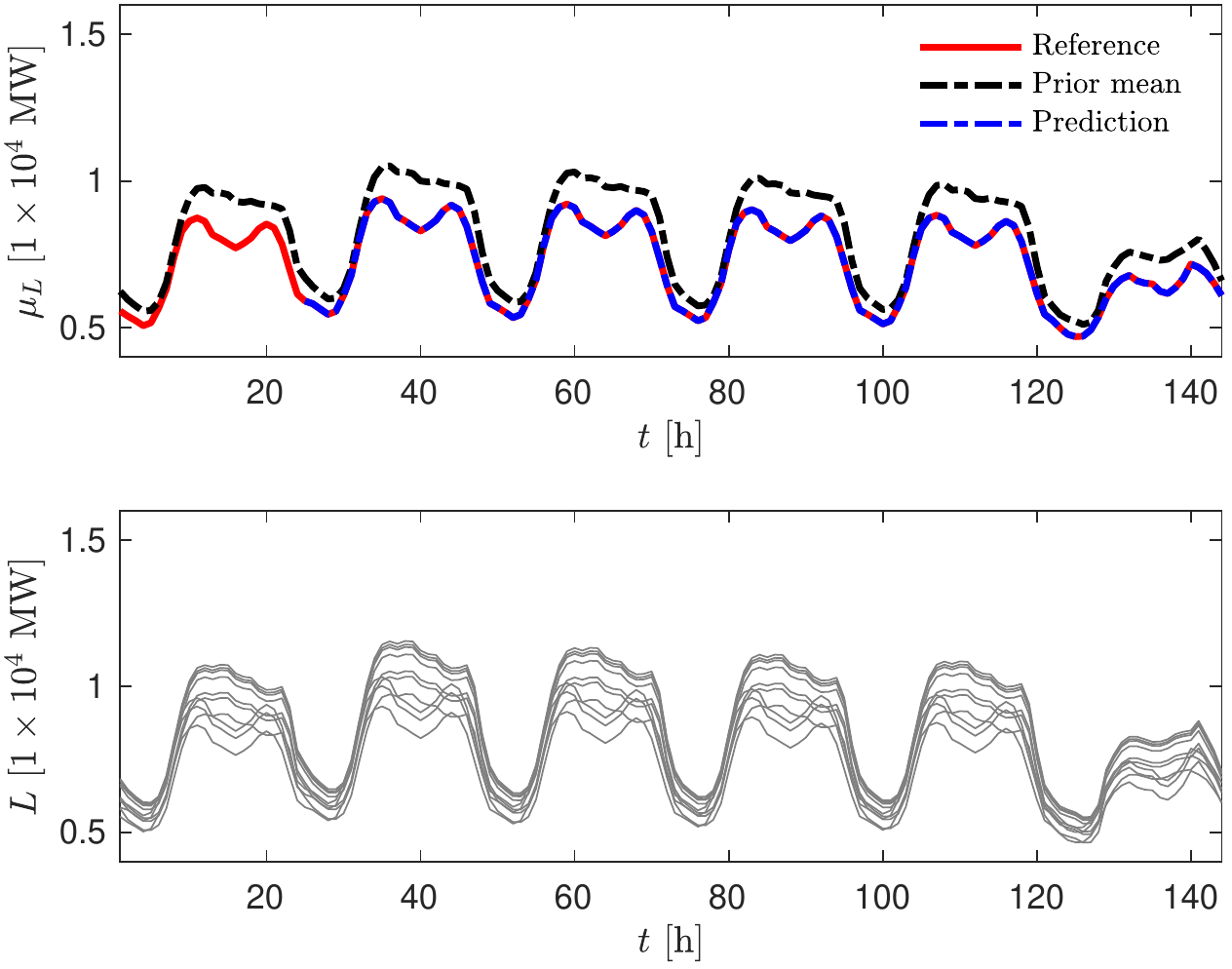}}\\
  \subfloat[10/13--10/19]{\includegraphics[width=0.45\textwidth]{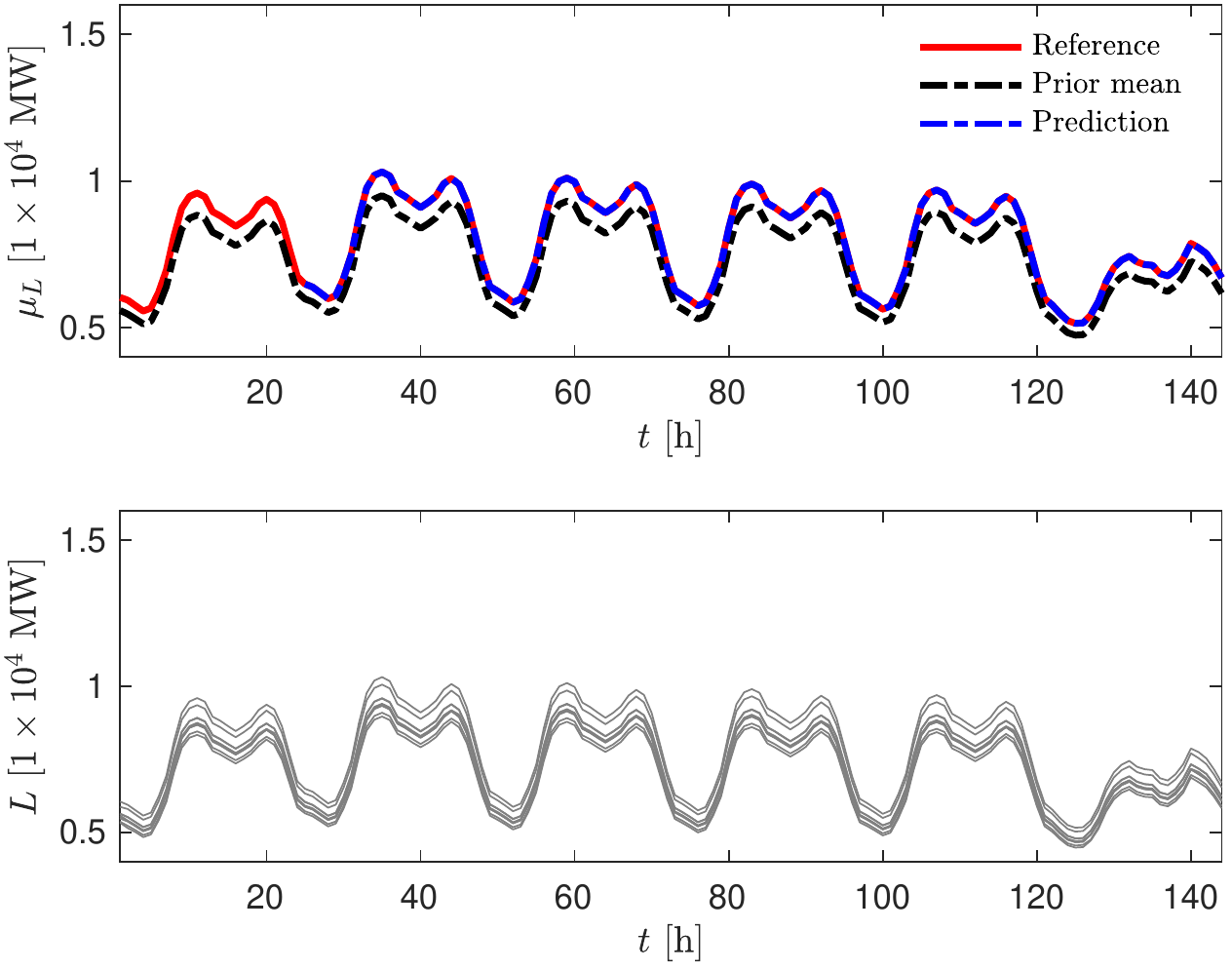}}\ %
  \subfloat[12/15--12/21]{\includegraphics[width=0.45\textwidth]{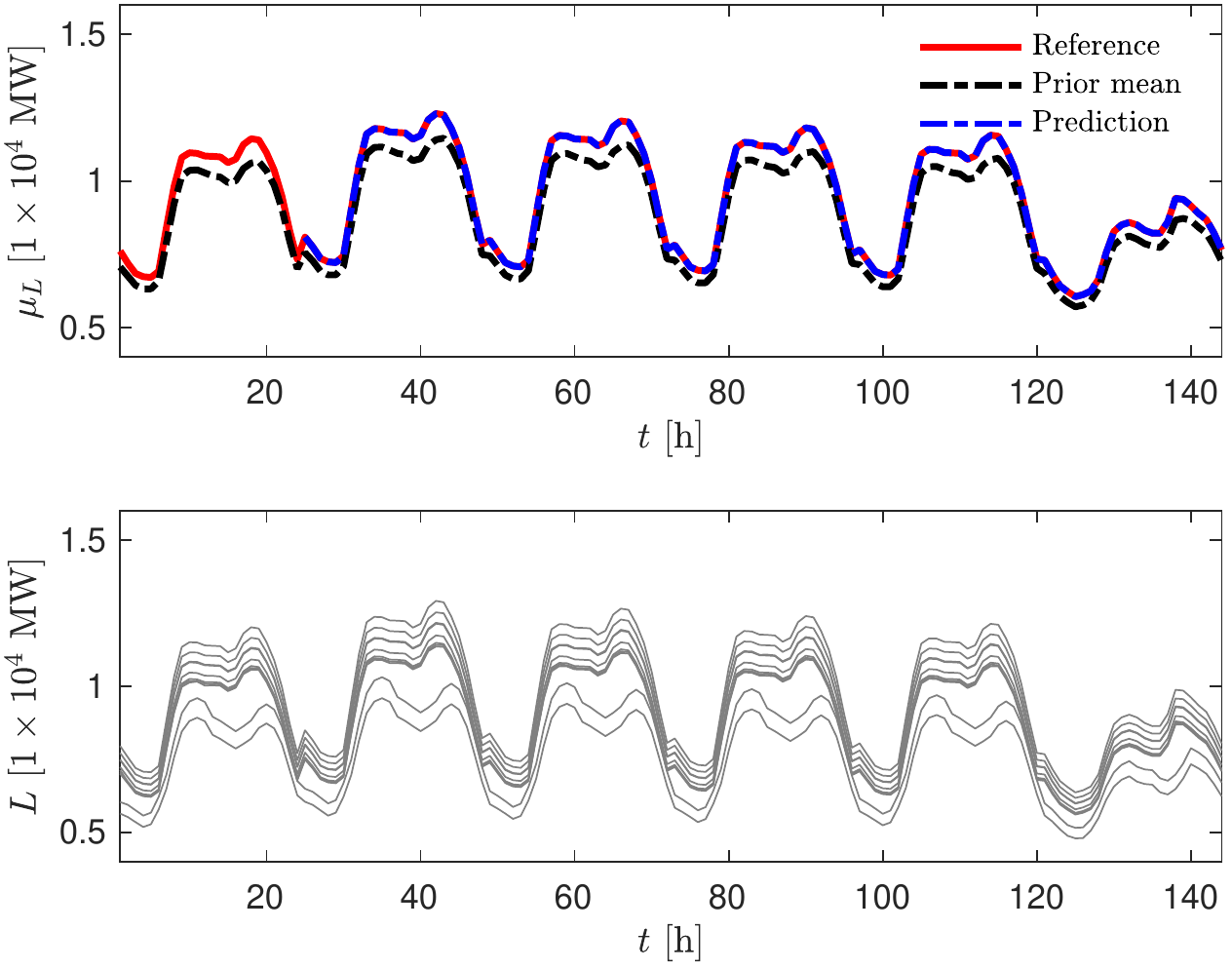}}%
  \caption{Weekly forecasts of total load for four weeks using 24h observations on Tuesday.
    Top panel: Prediction (blue) compared against reference (red) and the prior mean (black).
    Bottom panel: Ensemble of $10$ timeseries used to compute the empirical covariance.}
  \label{fig:load-forecast-wom}
\end{figure*}

\begin{figure*}[tbh]
  \centering%
  \subfloat[6/9--6/15]{\includegraphics[width=0.45\textwidth]{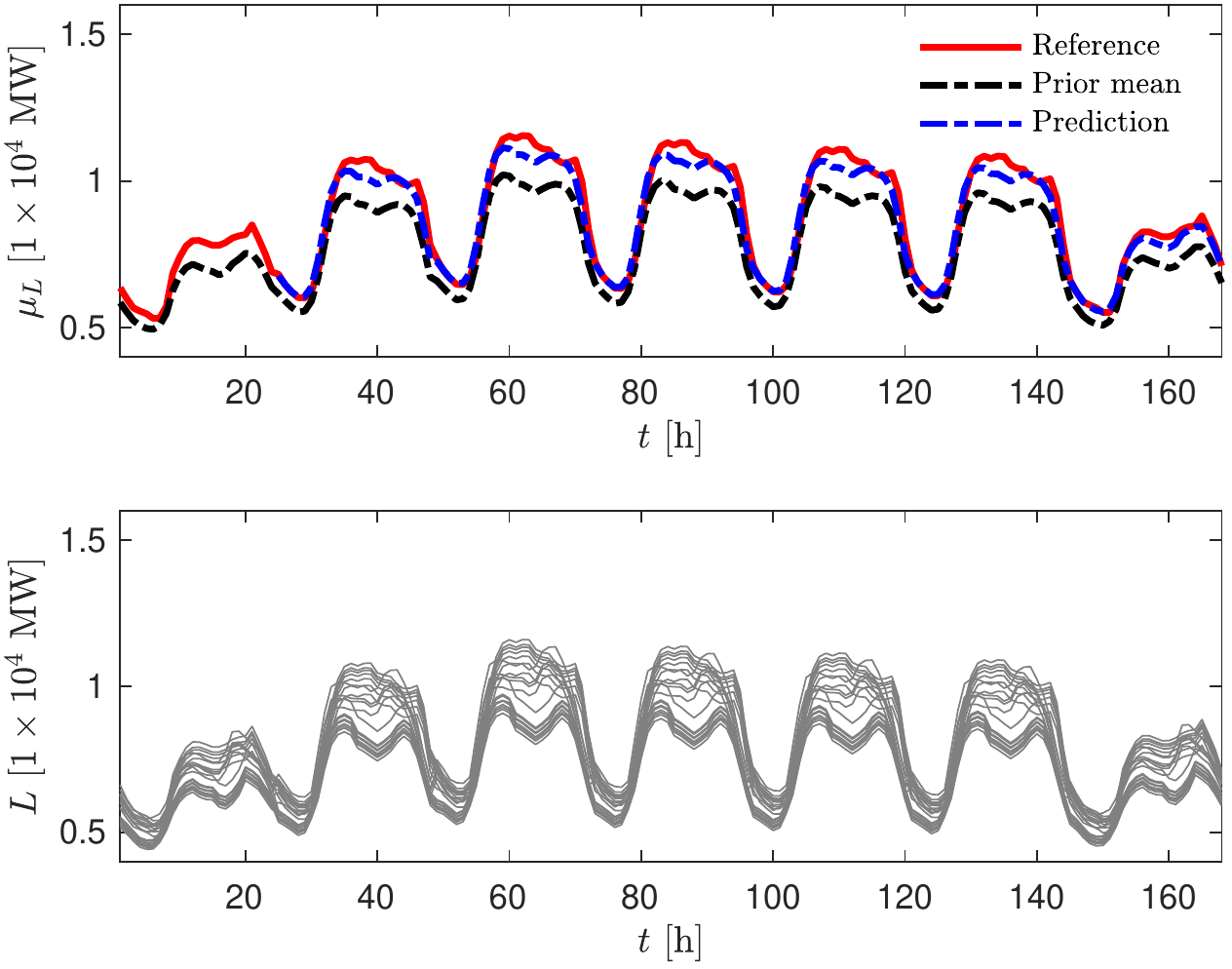}}\ %
  \subfloat[8/11--8/17]{\includegraphics[width=0.45\textwidth]{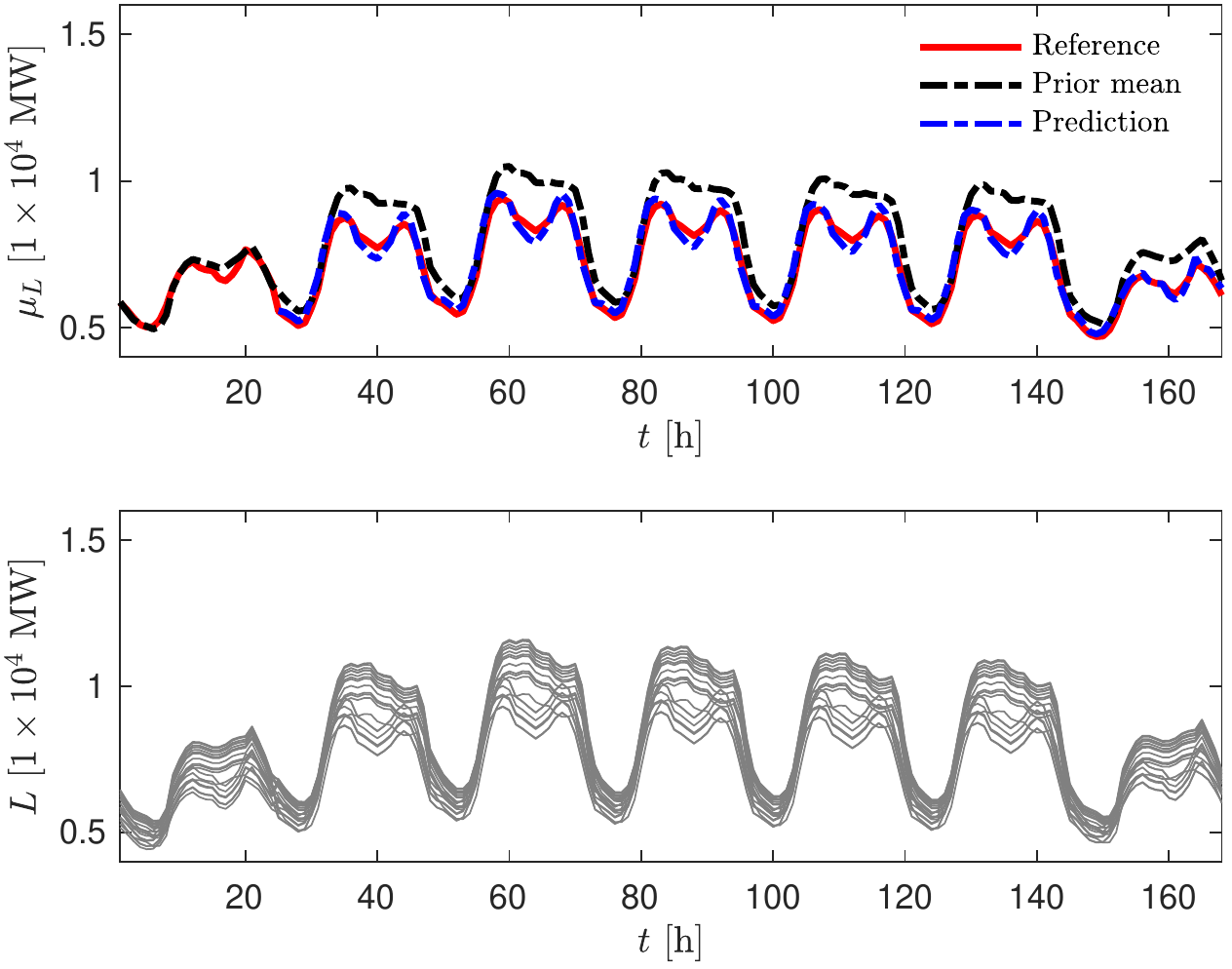}}\\
  \subfloat[10/13--10/19]{\includegraphics[width=0.45\textwidth]{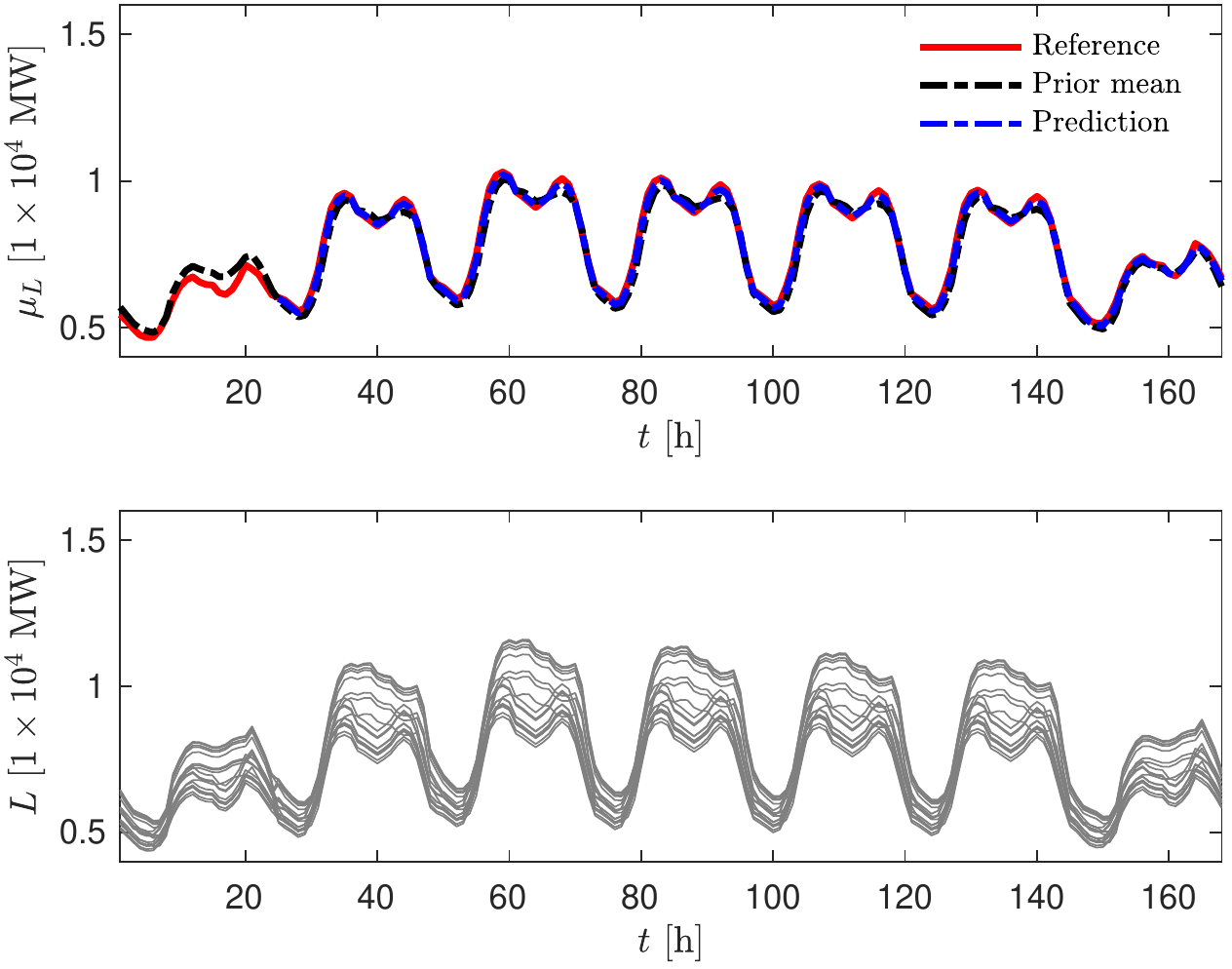}}\ %
  \subfloat[12/15--12/21]{\includegraphics[width=0.45\textwidth]{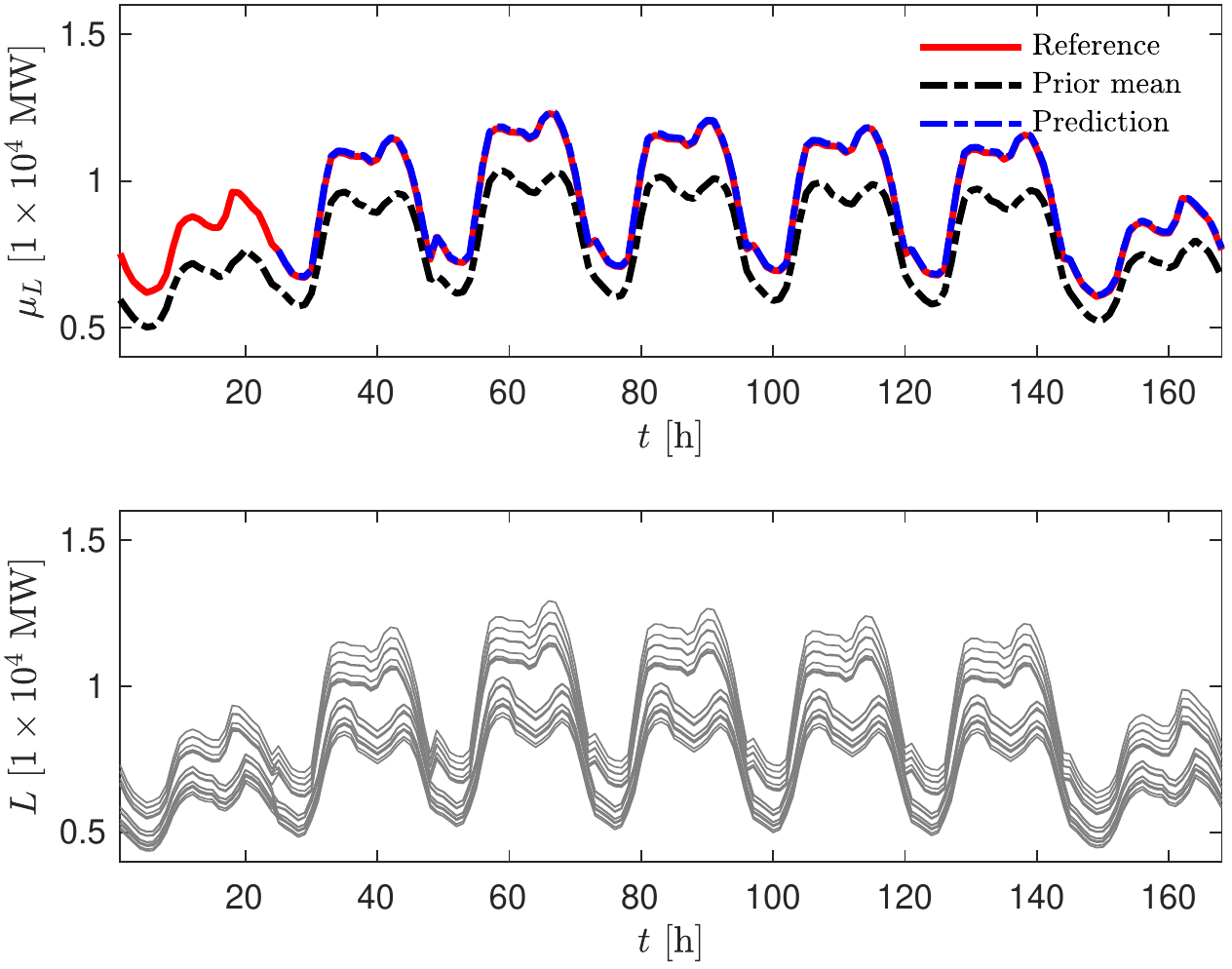}}%
  \caption{Weekly forecasts of total load for four weeks using 24h observations on Monday.
    Top panel: Prediction (blue) compared against reference (red) and the prior mean (black).
    Bottom panel: Ensemble of $20$ timeseries used to compute the empirical covariance.}
  \label{fig:load-forecast-wm}
\end{figure*}

\subsection{Weekly forecasting of total load demand}
\label{ss4}

\begin{figure}[htb]
  \centering%
  \includegraphics[width=0.9\linewidth]{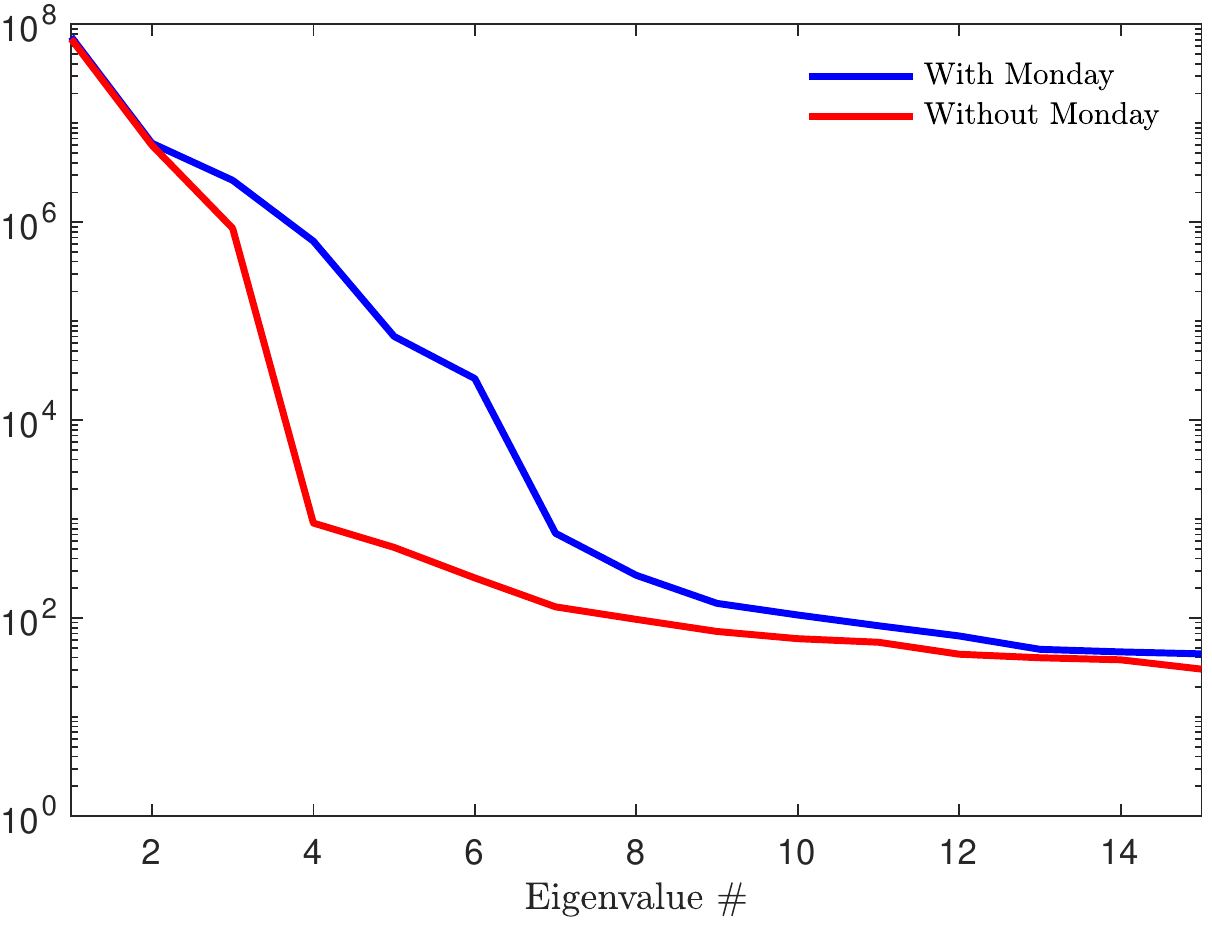}
  \caption{Eigenspectrum of empirical covariance for the week of 6/9--6/15 computed using $20$ realizations, with and without covariance}
  \label{fig:load-forecast-covar-spectrum}
\end{figure}

\begin{figure}[htb]
  \centering%
  \includegraphics[width=\linewidth]{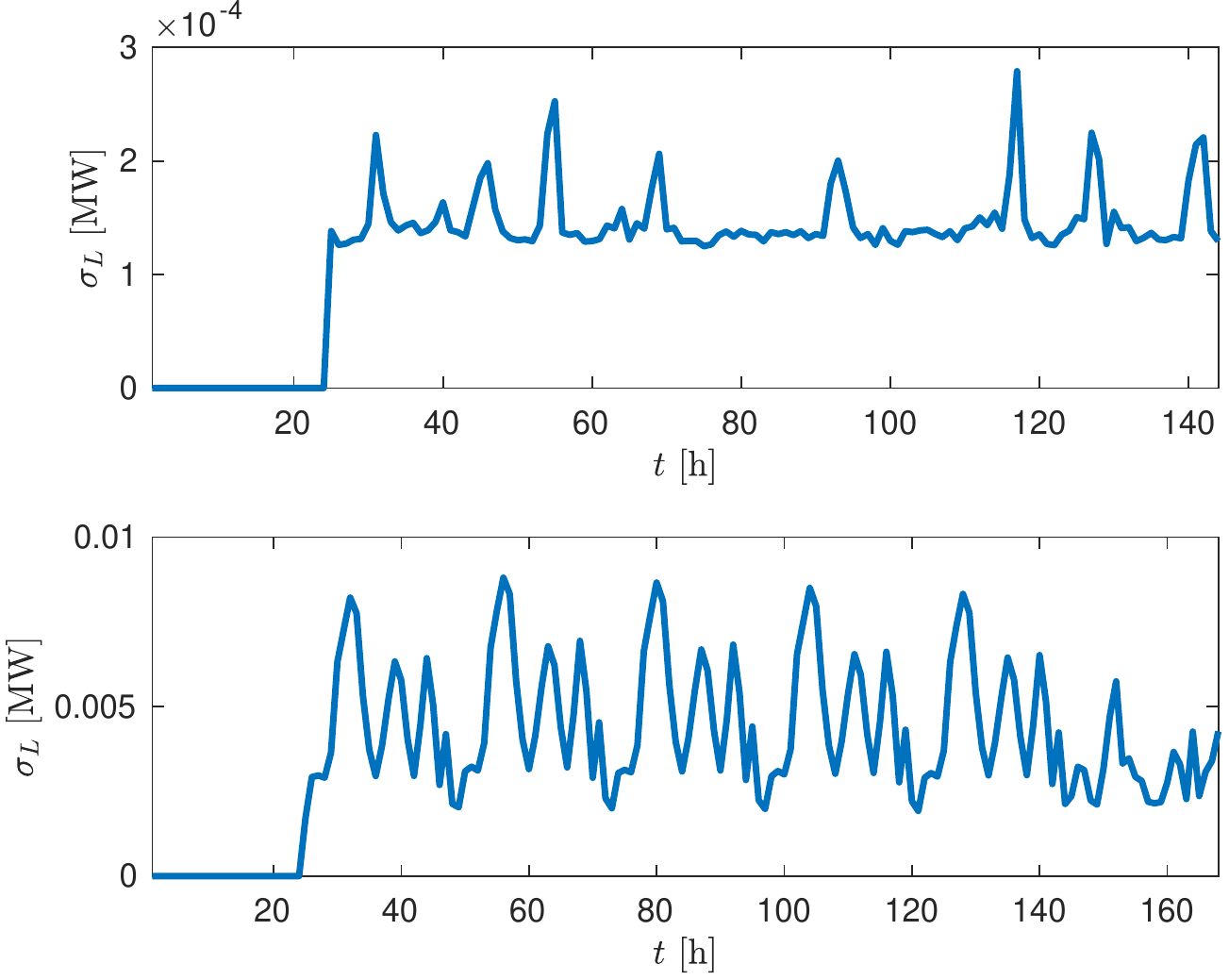}
  \caption{Posterior standard deviation of the weekly forecast for the week of 6/9--6/15.
    Top: Using 24h observations on Tuesday.
    Bottom: Using 24h observations on Monday.}
  \label{fig:load-forecast-std}
\end{figure}

\subsubsection{Ensemble GPR}

To employ the proposed EGPR method we split the historic data into weekly timeseries.
The EGPR forecasts for a given week are computed by constructing an ensemble of the $N$ previous weeks.
As reference we employ the data for the week to be forecasted.
An analysis of the data reveals that in the short term, Monday total load demand is weakly correlated to the total load demand of the other days of the week.
This can be seen in Figure~\ref{fig:load-forecast-covar}, where we show the ensemble covariance for the forecast of 6/9--6/15, computed using $N = 5$ and $N = 20$.
In the short term ($N = 5$), it can be seen that Mondays are weakly correlated with the rest of the weekdays, while other weekdays are strongly correlated with each other.
We find that for the considered data, the correlations between Mondays and the rest of the weekdays are resolved by the ensembles by taking $N = 20$ or longer.
We note that the correlations between days of the week are artifacts of the synthetic data generation process outlined in Section~\ref{sec:synthetic} and don't result from the EGPR method.

This analysis suggests that two forecasts can be performed: (i) weekly forecasting using 24h data on Tuesday, and (ii) weekly forecasting using 24h data on Monday with, each with different values of $N$ so that the correlations between the days of the week are sufficiently resolved.
These forecasts are presented for the considered weeks in Figures~\ref{fig:load-forecast-wom} and \ref{fig:load-forecast-wm}, respectively.
Figure~\ref{fig:load-forecast-wom} shows that the weekly forecasting using 24h data on Tuesday accurately resolves the reference total load demand.
Most importantly, the forecast does not collapse into the prior mean, which is a common pathology of GPR-based forecasting.
Figure~\ref{fig:load-forecast-wm} shows that weekly forecasting using 24h on Monday is less accurate, specially for the weeks of 6/9--6/15 and 8/11--8/17.
Nevertheless, the pathology of collapse towards the prior mean is again avoided.
This is because, as seen in Figure~\ref{fig:load-forecast-covar}, even with $N = 20$, Mondays are more weakly correlated to the other days of the week than the other days of the week with each other.

In order to study the accuracy of the proposed EGPR, we consider the eigenspectrum (i.e., the set of eigenvalues) of the empirical covariance.
Figure~\ref{fig:load-forecast-covar-spectrum} shows the eigenspectrums of the empirical covariance with $N = 20$ employed for forecasting the week of 6/9--6/15.
Following the previous discussion, we compute the eigenspectrum by first keeping Monday data in the ensemble, and then by excluding Monday data in the ensemble.
It can be seen that the eigenvalues of the empirical covariance excluding Mondays decay faster than those of the empirical covariance including Mondays.
This is in accordance with Figure~\ref{fig:load-forecast-covar}, which shows that excluding Mondays results in a more structured weekly covariance pattern.
As a consequence, excluding Mondays results in a EGPR model with lower effective stochastic dimension, which allows us to both use less ensemble members to compute the EGPR prior covariance and to use less data for forecasting.
Therefore, it can be expected that for the same amount of data, forecasting excluding Mondays will result in forecasts with tighter credibility bounds than by including Mondays.
This is verified by Figure \ref{fig:load-forecast-std}, which shows the posterior standard deviation for the forecast of the week of 6/9--6/15.
It can be seen that forecasting using 24h data on Tuesday results in lower posterior standard deviation than by using 24h data on Monday, and in accordance with Figures~\ref{fig:load-forecast-wom} and \ref{fig:load-forecast-wm}, more accurate forecasts.
We note that the low-dimensional structure of the EGPR model stems directly from the low-dimensional structure of the data, which in turn stems from the synthetic data generation procedure (Section~\ref{sec:synthetic}).

\subsubsection{Standard data-driven GPR}

\begin{figure}[htb]
    \centering
	\includegraphics[scale=0.32]{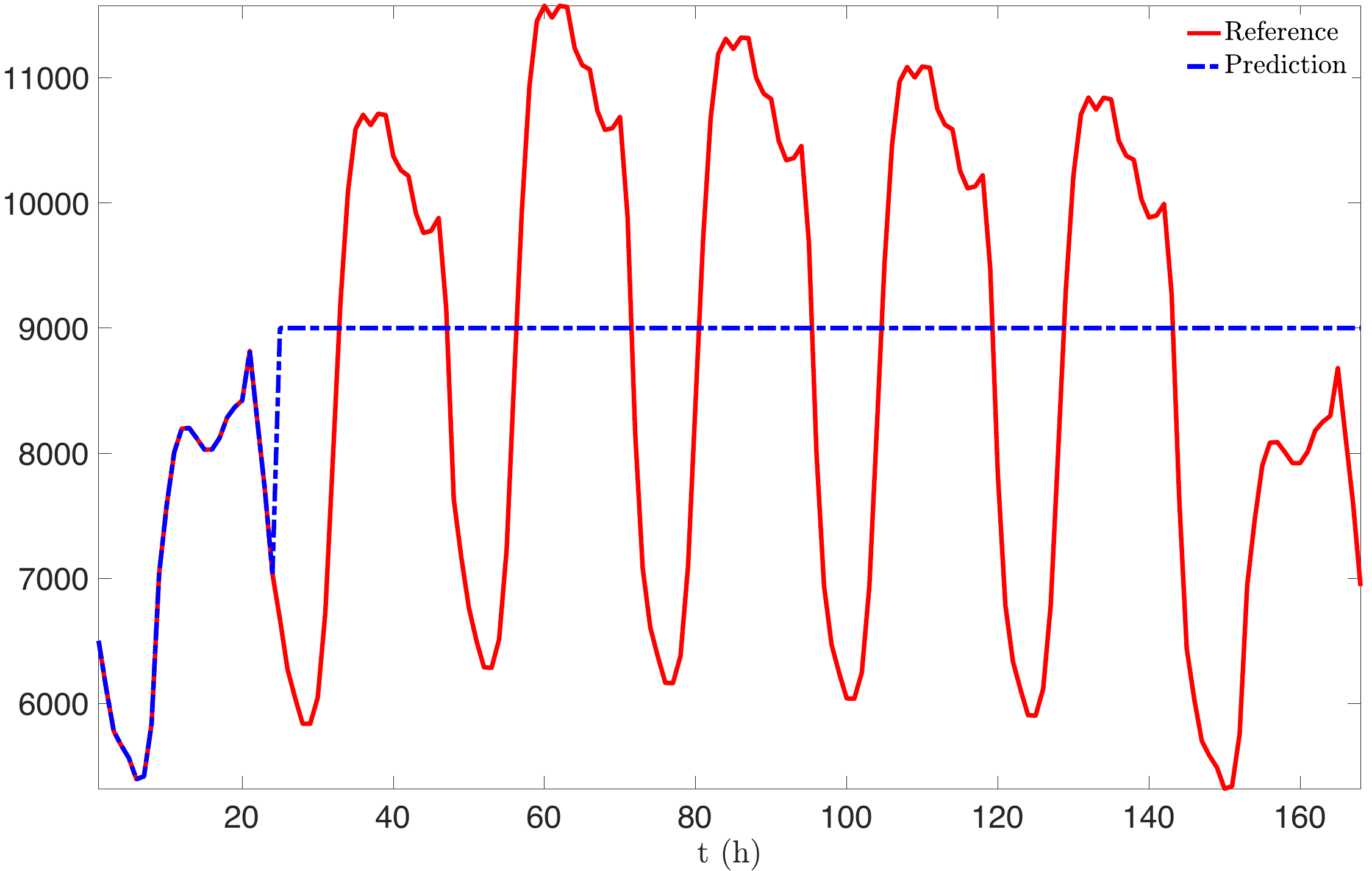}
	\caption{Weekly total load forecasting using standard GPR for 06/23-06/29 (Mon-Sun).}
	\label{g18}       
\end{figure}
\begin{figure}[htb]
    \centering
	\includegraphics[scale=0.32]{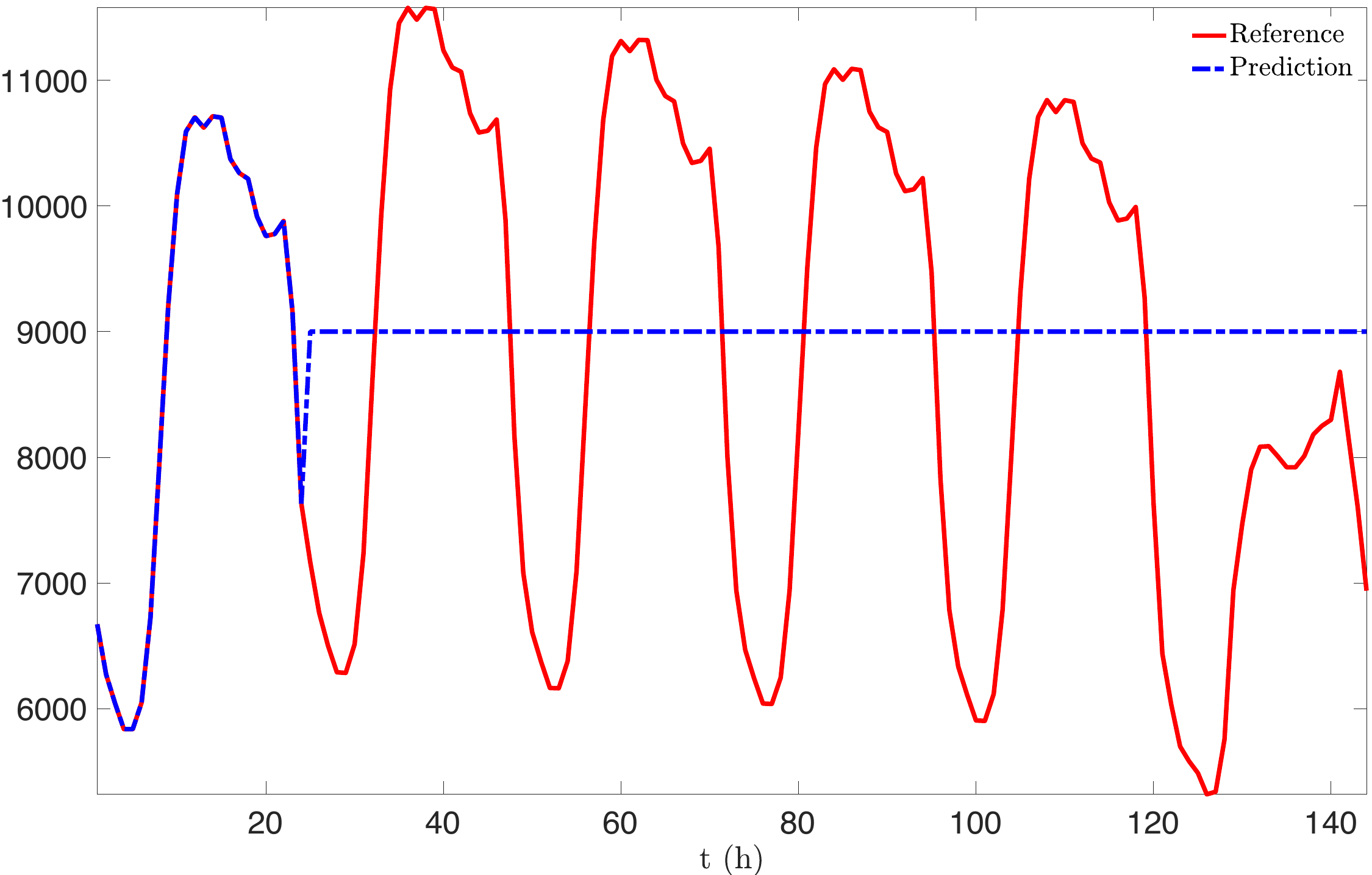}
	\caption{Weekly total load forecasting using standard GPR for 06/24-06/29 (Tues-Sun).}
	\label{g17}       
\end{figure}

Figures~\ref{g18} and~\ref{g17} show the standard  GPR forecast of total load for 06/23-06/29 (Mon-Sun) and 06/24-06/29 (Tues-Sun), correspondingly. Here we assume the ``Squared Exponential'' (SE) form for the prior covariance functions $K_{oo}$, $K_{of}$, and $K_{ff}$,
\begin{equation*}
    K(t_1,t_2) = \sigma^2 \exp \left ( -\frac{|t_1 - t_2|}{2 \gamma^2} \right ),
\end{equation*}
parameterized by the standard deviation $\sigma$ and the correlation time $\gamma$.
The prior parameters $\overline{L^o}$, $\overline{L^f}$, $\sigma$ and $\gamma$ by  maximizing the marginal likelihood of the load data observed throughout the previous week of 06/16-06/22.
As we can see from Figures~\ref{g18} and~\ref{g17}, the standard GPR drives the forecast to an average value of the load obtained from the previous week (i.e., collapse towards the prior mean).
We can see that the proposed EGPR method significantly outperforms the standard GPR with SE covariance. 

\subsubsection{ARIMA}

\begin{figure}[htb]
    \centering
	\includegraphics[scale=0.32]{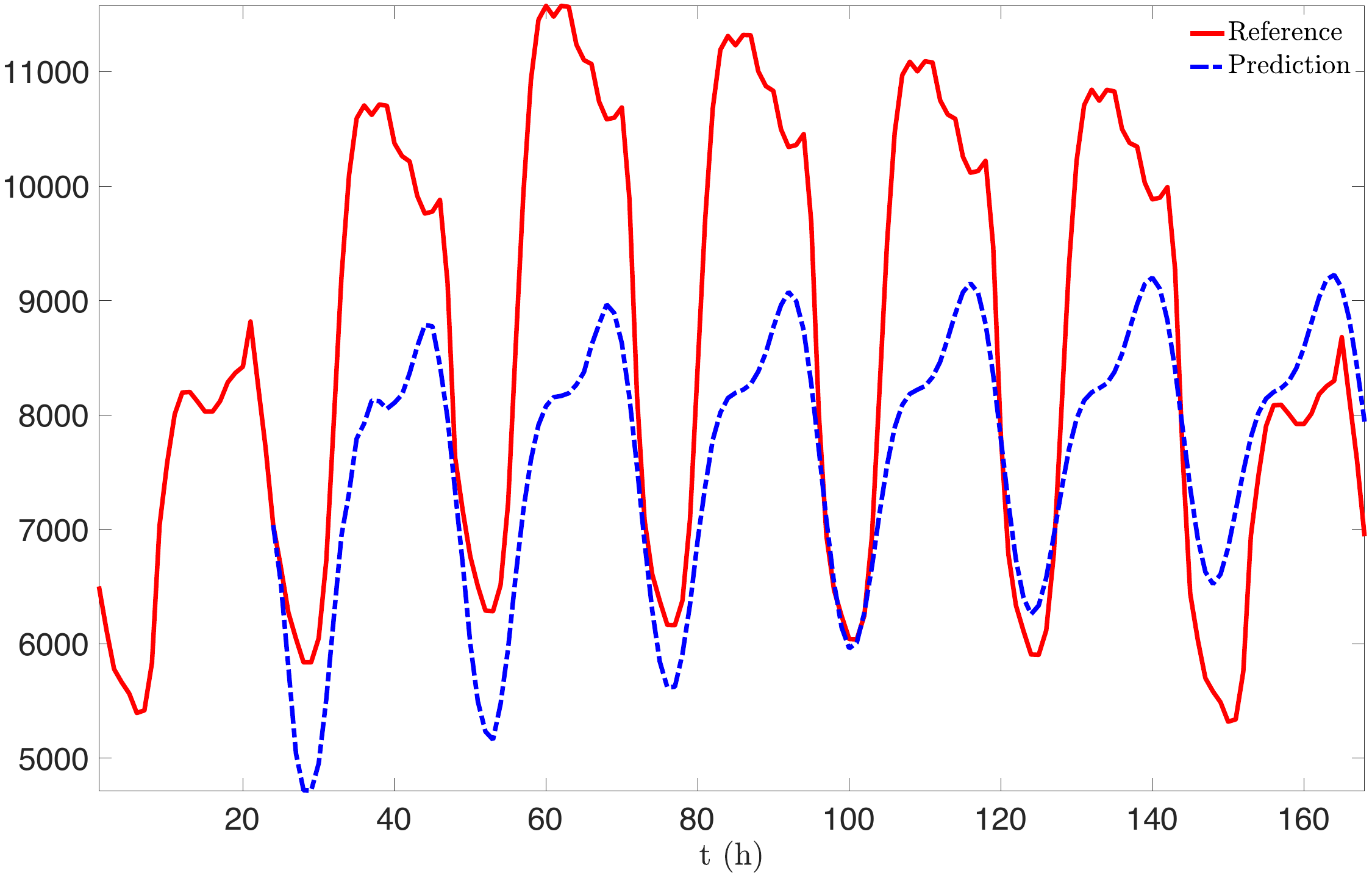}
	\caption{Weekly total load forecasting using ARIMA for 06/23-06/29 (Mon-Sun).}
	\label{g21}       
\end{figure}
\begin{figure}[htb]
    \centering
	\includegraphics[scale=0.32]{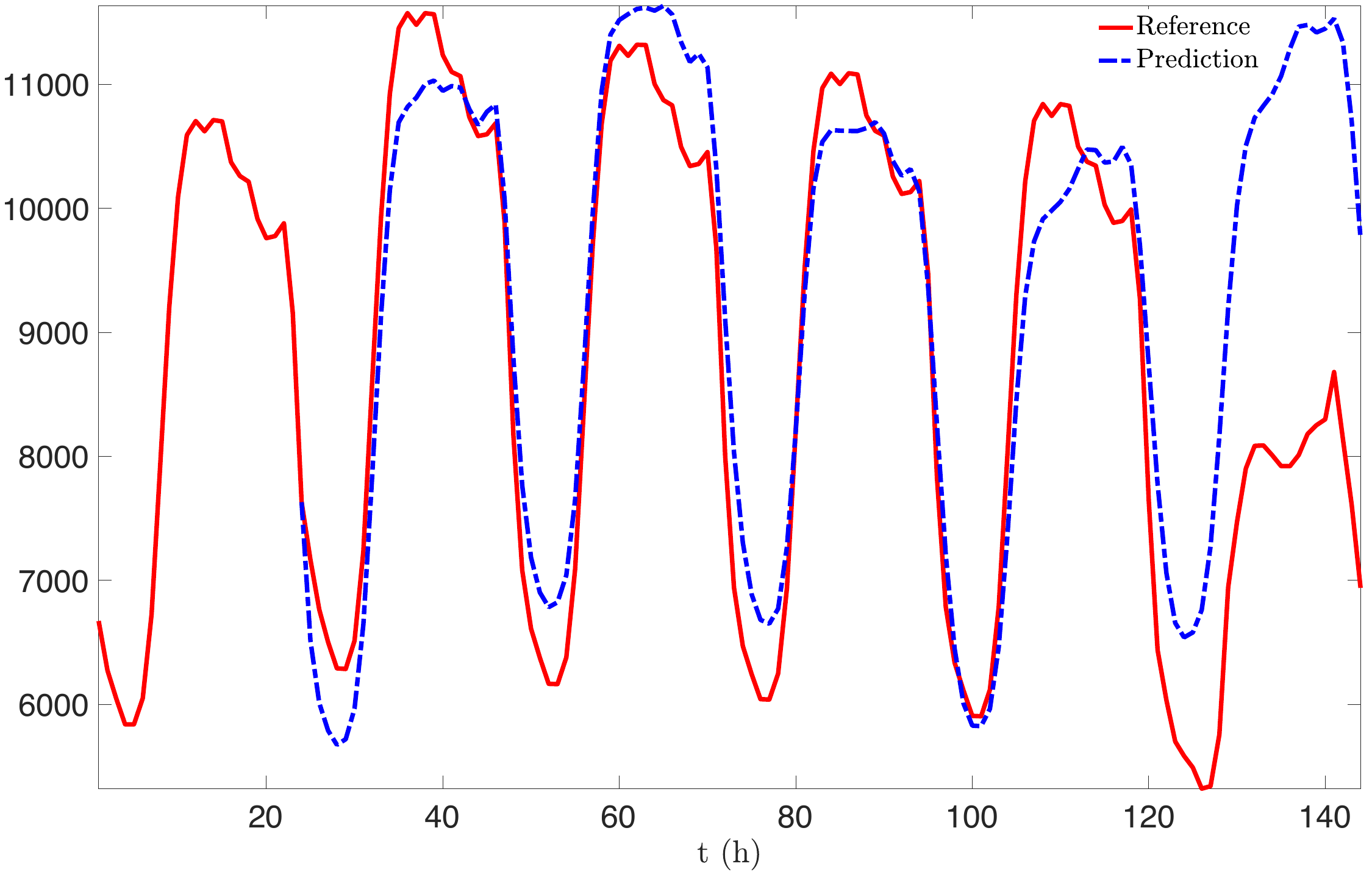}
	\caption{Weekly total load forecasting using ARIMA for 06/24-06/29 (Tues-Sun).}
	\label{g22}       
\end{figure}

In this section we employ the nonseasonal ARIMA method 
\cite{hipel} to forecast total load demand.
Figures~\ref{g21} and \ref{g22} show the ARIMA forecasts of total load for 06/23--6/29 (Mon-Sun) and 06/24--06/29 (Tues-Sun), respectively, using the observed load data from the previous week as training data.
For the six-day forecast shown in Figure~\ref{g21}, we use data from the previous seven days.
For the five day forecast shown in Figure~\ref{g22}, we use data from the previous six days.
The size of the training data plays an important role in determining the order of autoregressive model.
The order of the ARIMA model for Figures~\ref{g21} and \ref{g22} are 24 and 18, respectively.
It can be seen that ARIMA performs significantly better than the standard GPR method, but worse worse than the proposed EGPR method.

\begin{figure*}[thb]
  \centering%
  \subfloat[6/9--6/15]{\includegraphics[width=0.45\textwidth]{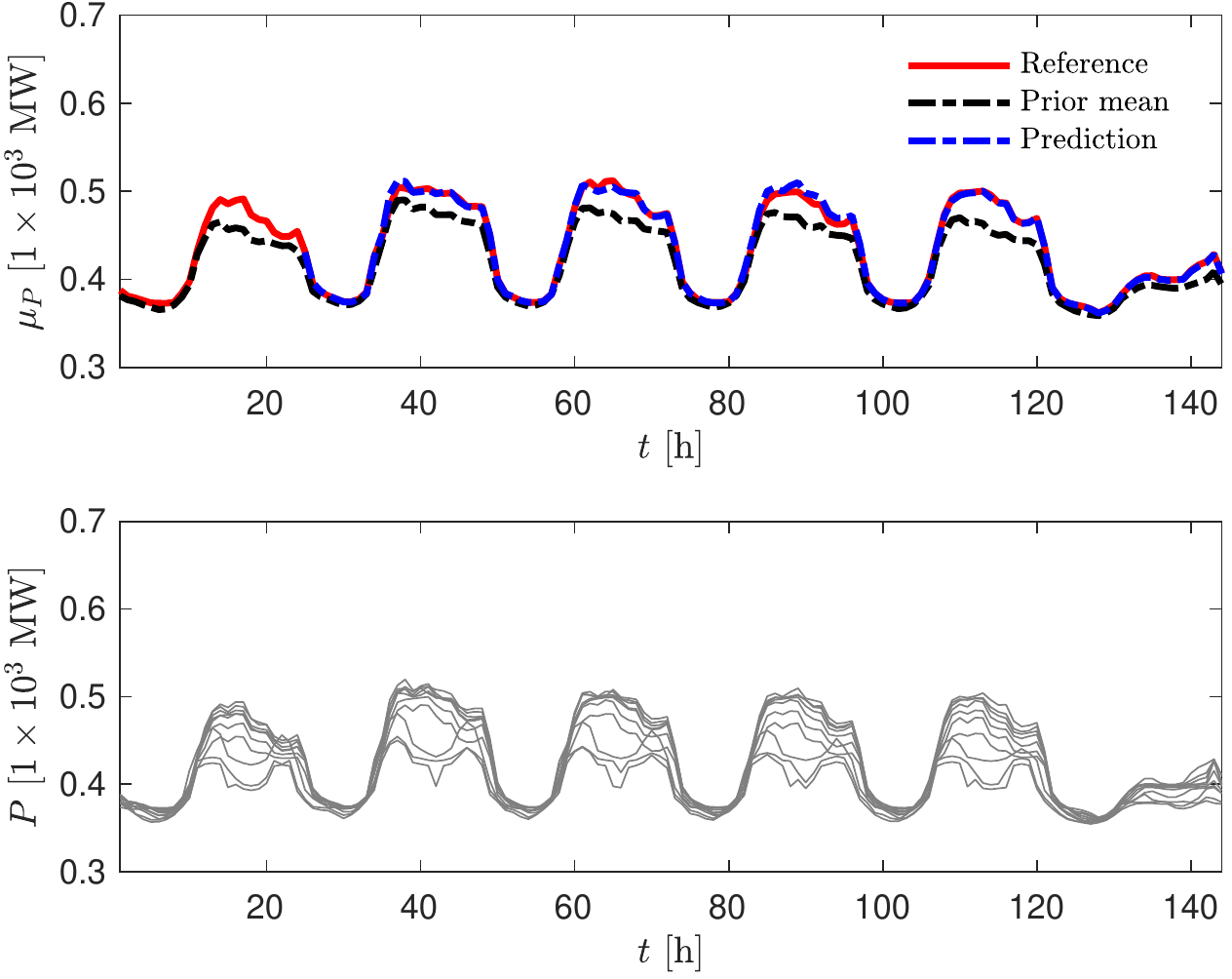}}\ %
  \subfloat[8/11--8/17]{\includegraphics[width=0.45\textwidth]{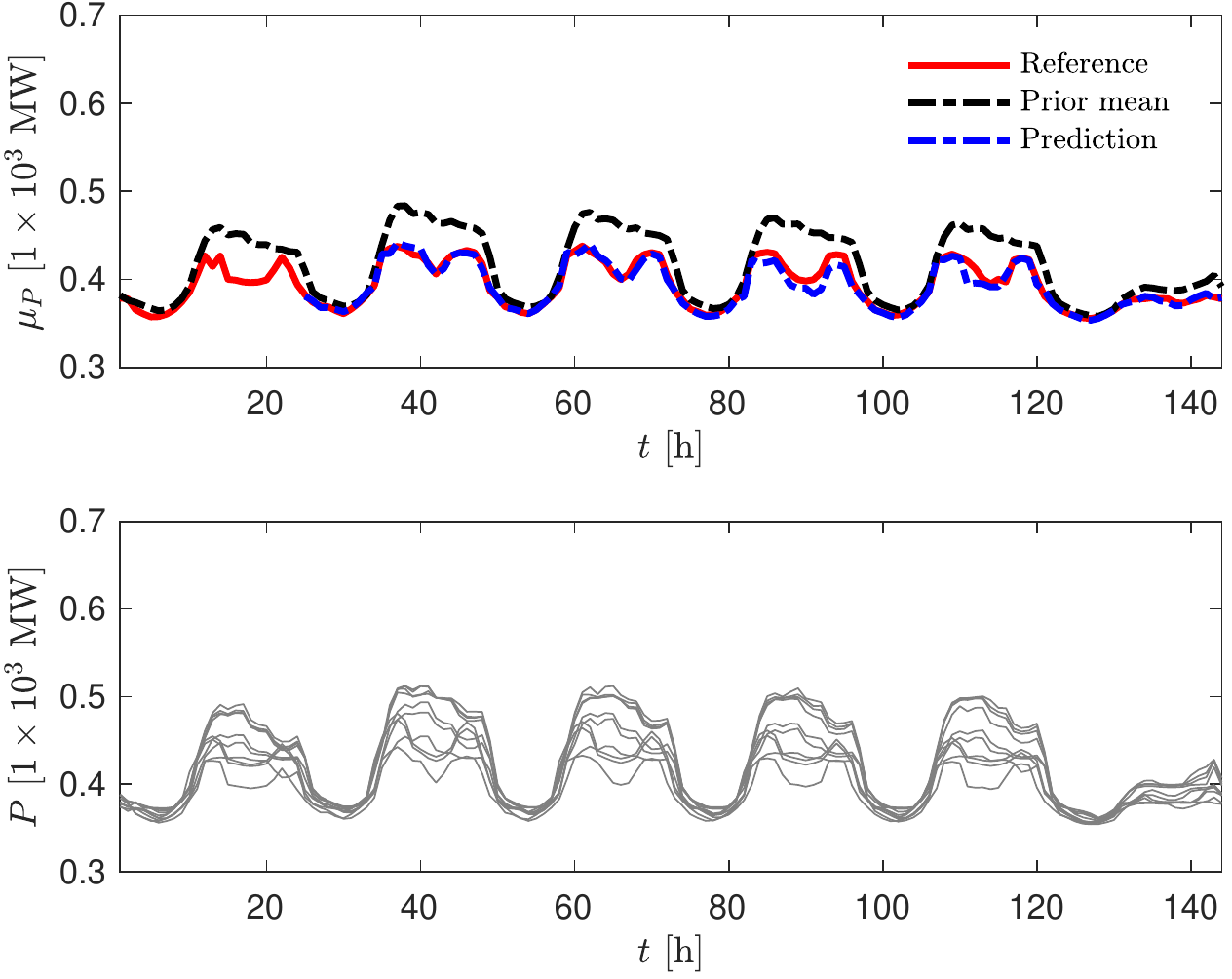}}\\
  \subfloat[10/13--10/19]{\includegraphics[width=0.45\textwidth]{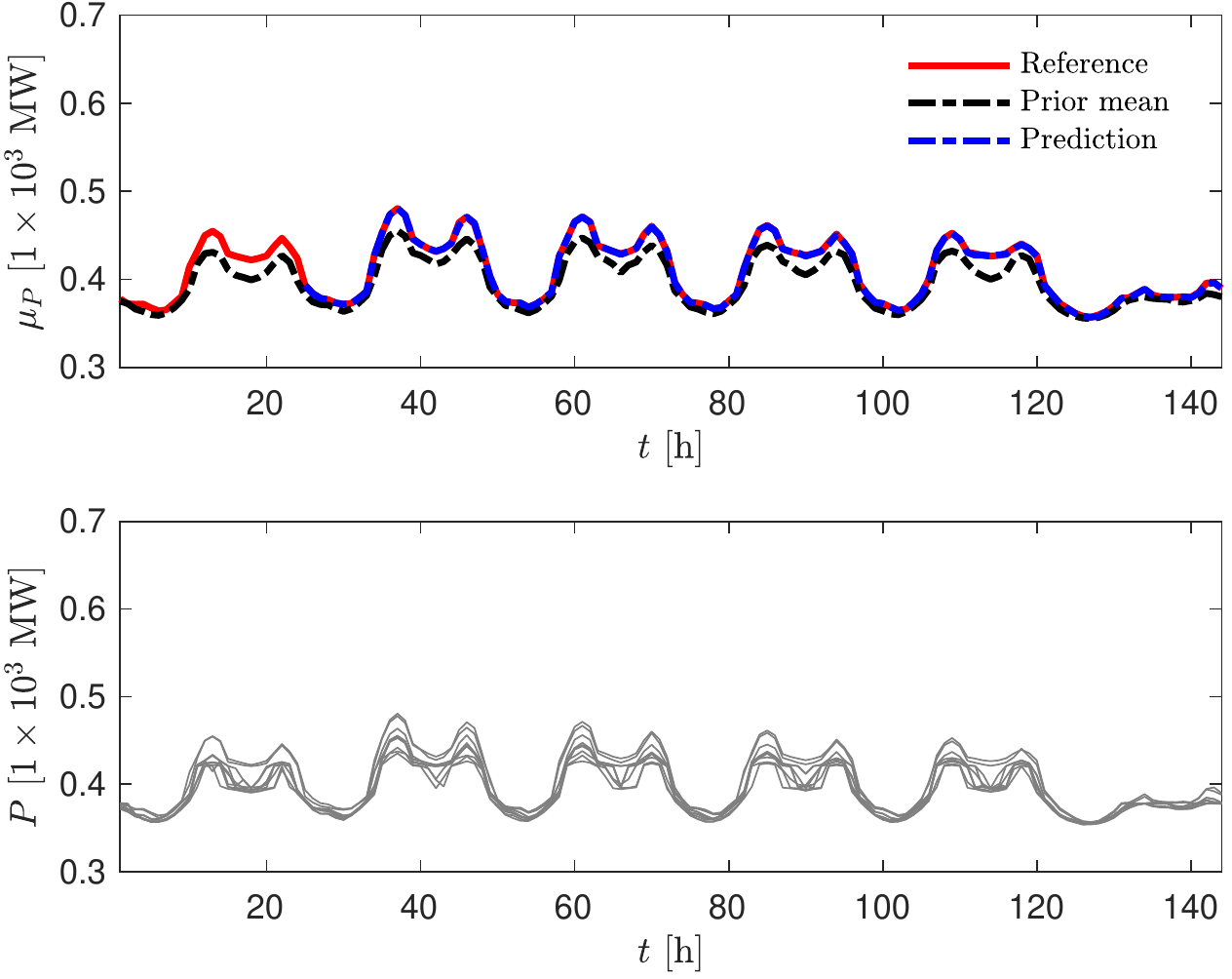}}\ %
  \subfloat[12/15--12/21]{\includegraphics[width=0.45\textwidth]{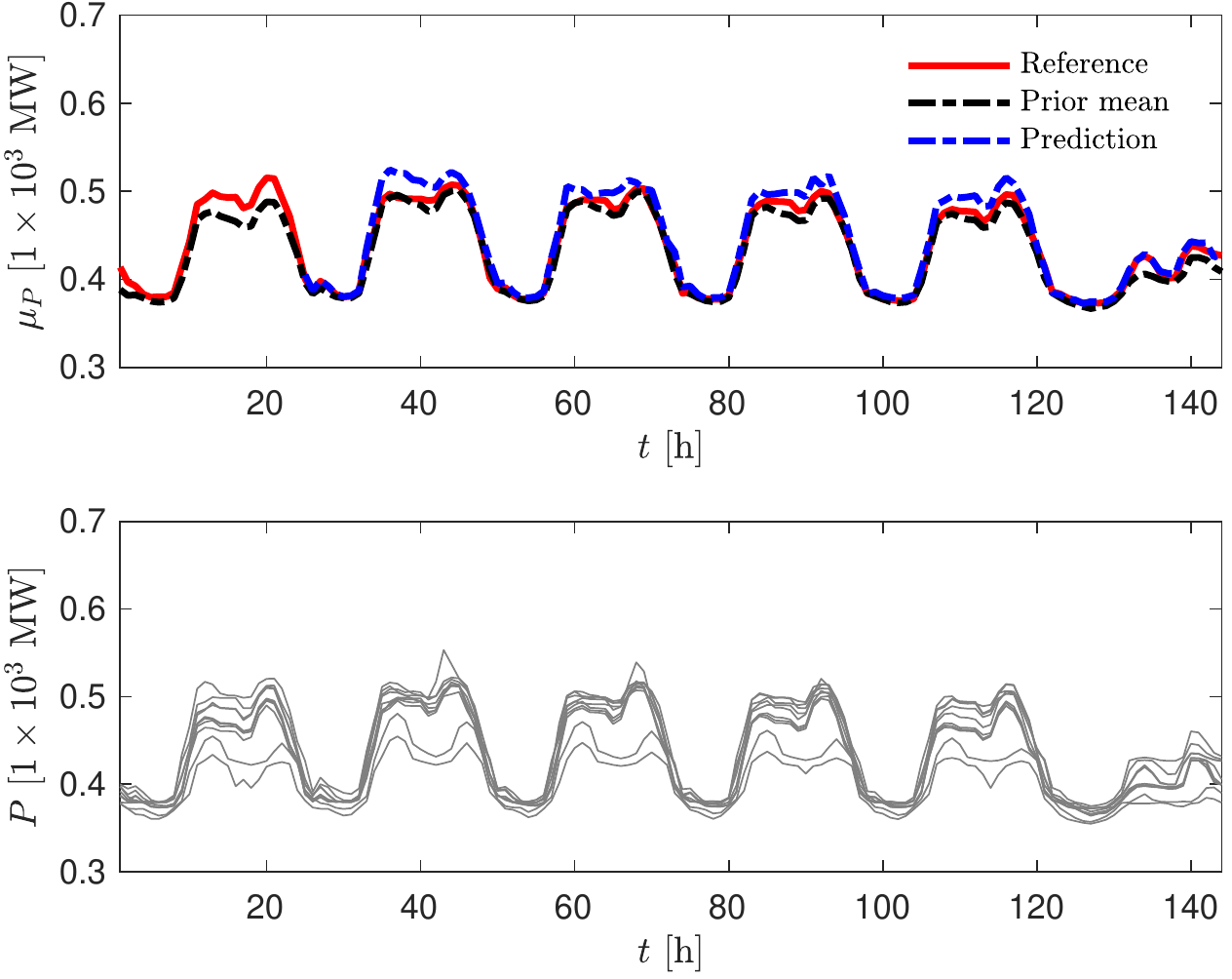}}%
  \caption{Weekly forecasts of the generation schedule for generator $15$ for four weeks using 24h observations on Tuesday.
    Top panel: Prediction (blue) compared against reference (red) and the prior mean (black).
    Bottom panel: Ensemble of $10$ timeseries used to compute the empirical covariance.}
  \label{fig:gen-forecast-wom} 
\end{figure*}

\begin{figure*}[thb]
  \centering%
  \subfloat[6/9--6/15]{\includegraphics[width=0.45\textwidth]{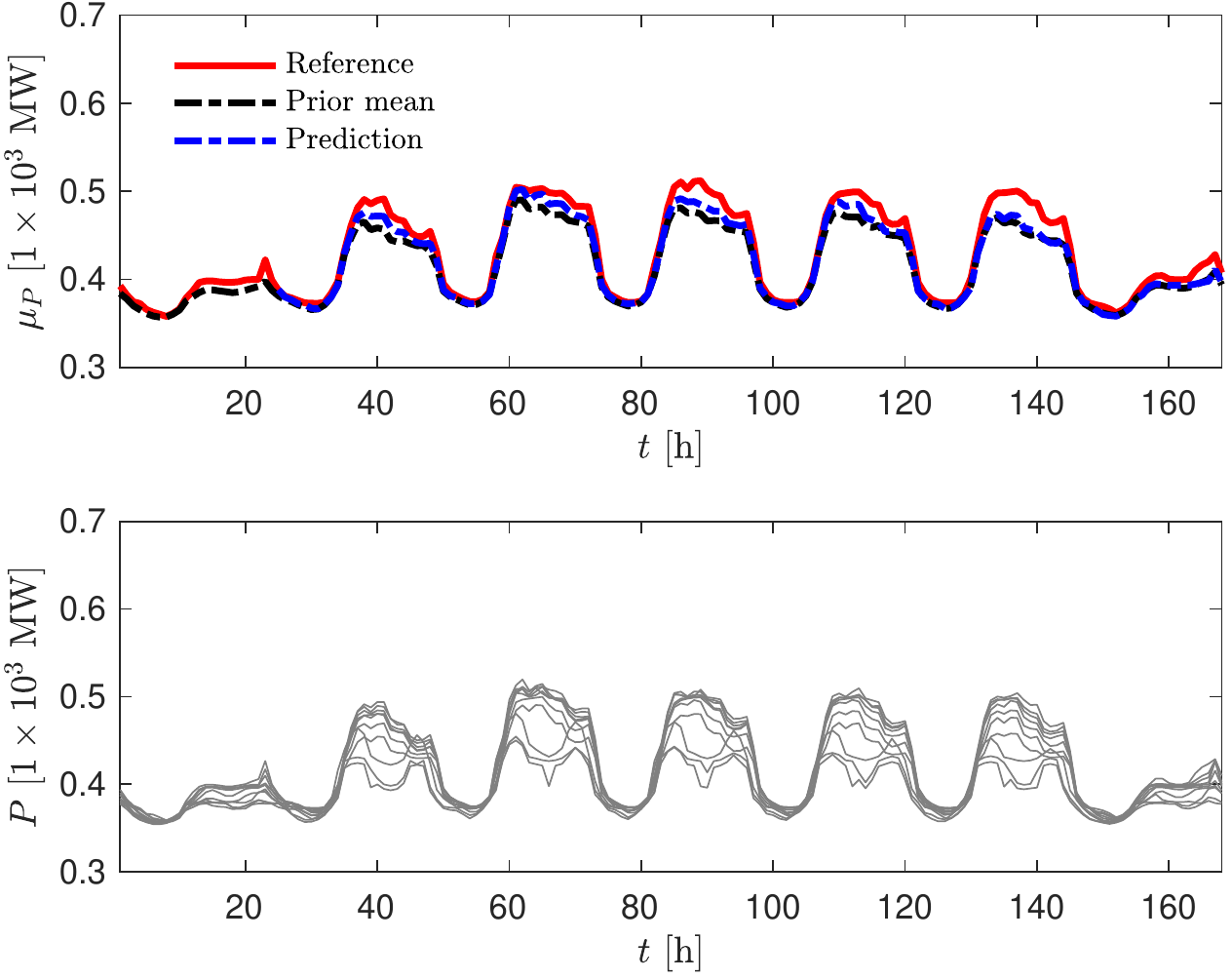}}\ %
  \subfloat[8/11--8/17]{\includegraphics[width=0.45\textwidth]{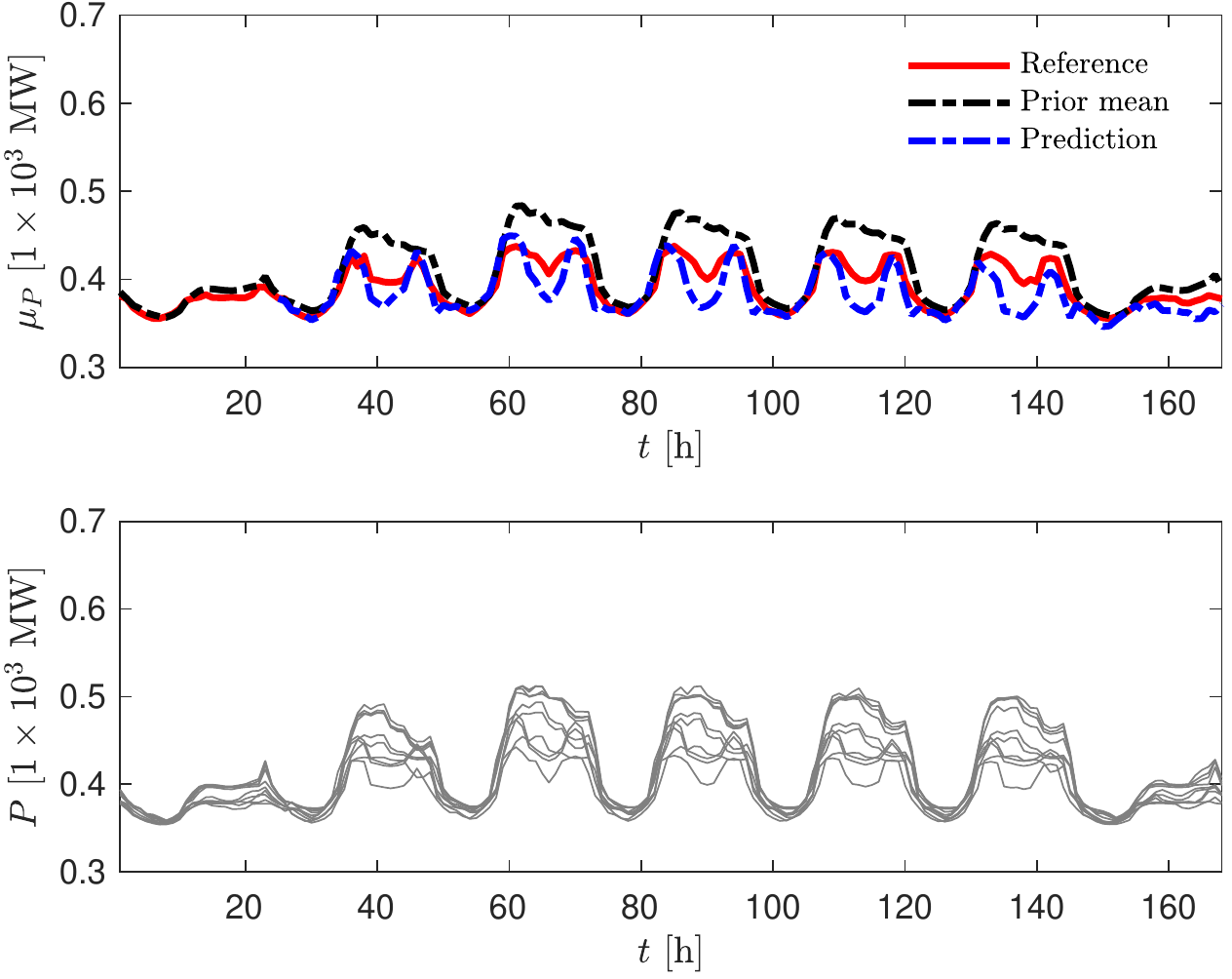}}\\
  \subfloat[10/13--10/19]{\includegraphics[width=0.45\textwidth]{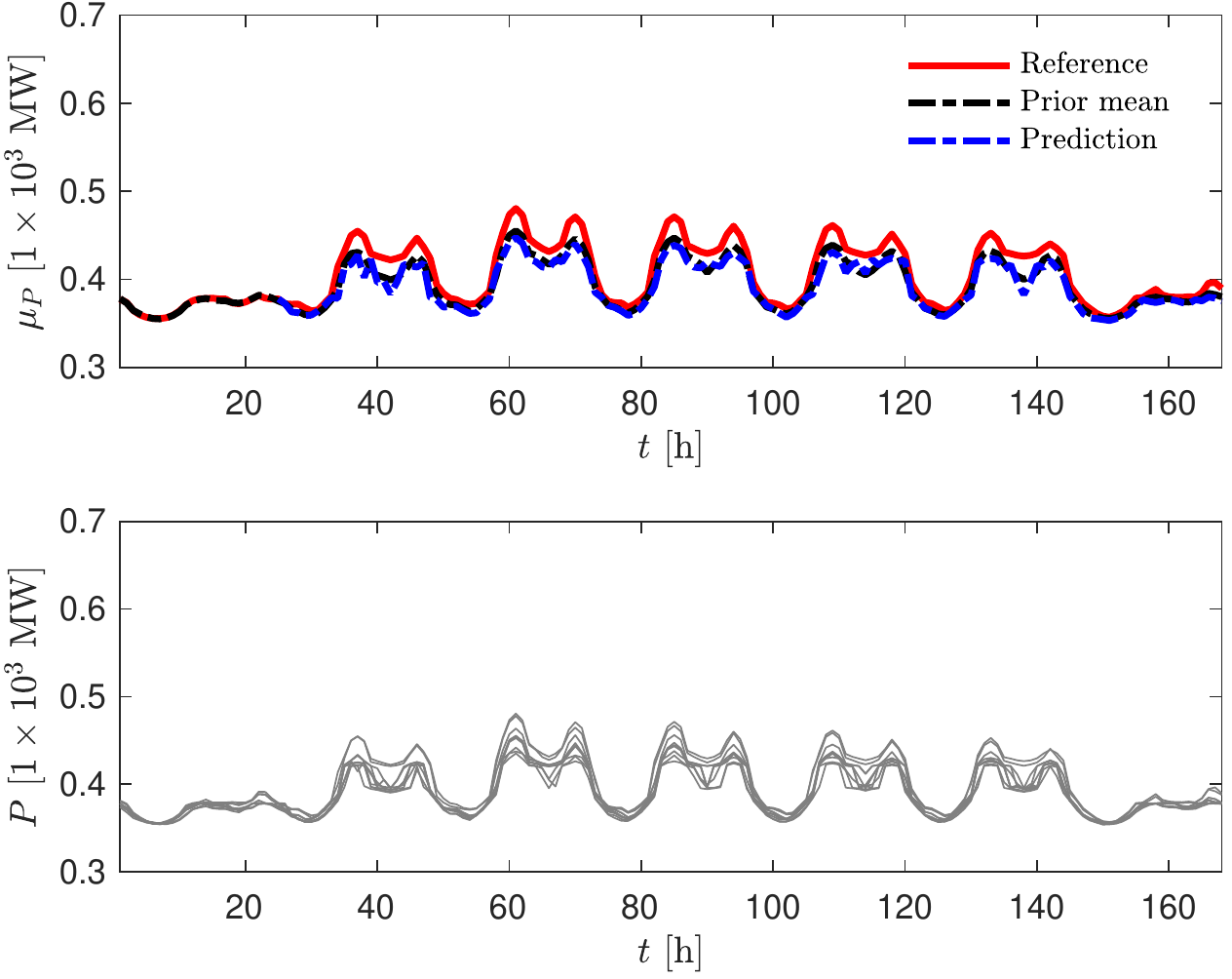}}\ %
  \subfloat[12/15--12/21]{\includegraphics[width=0.45\textwidth]{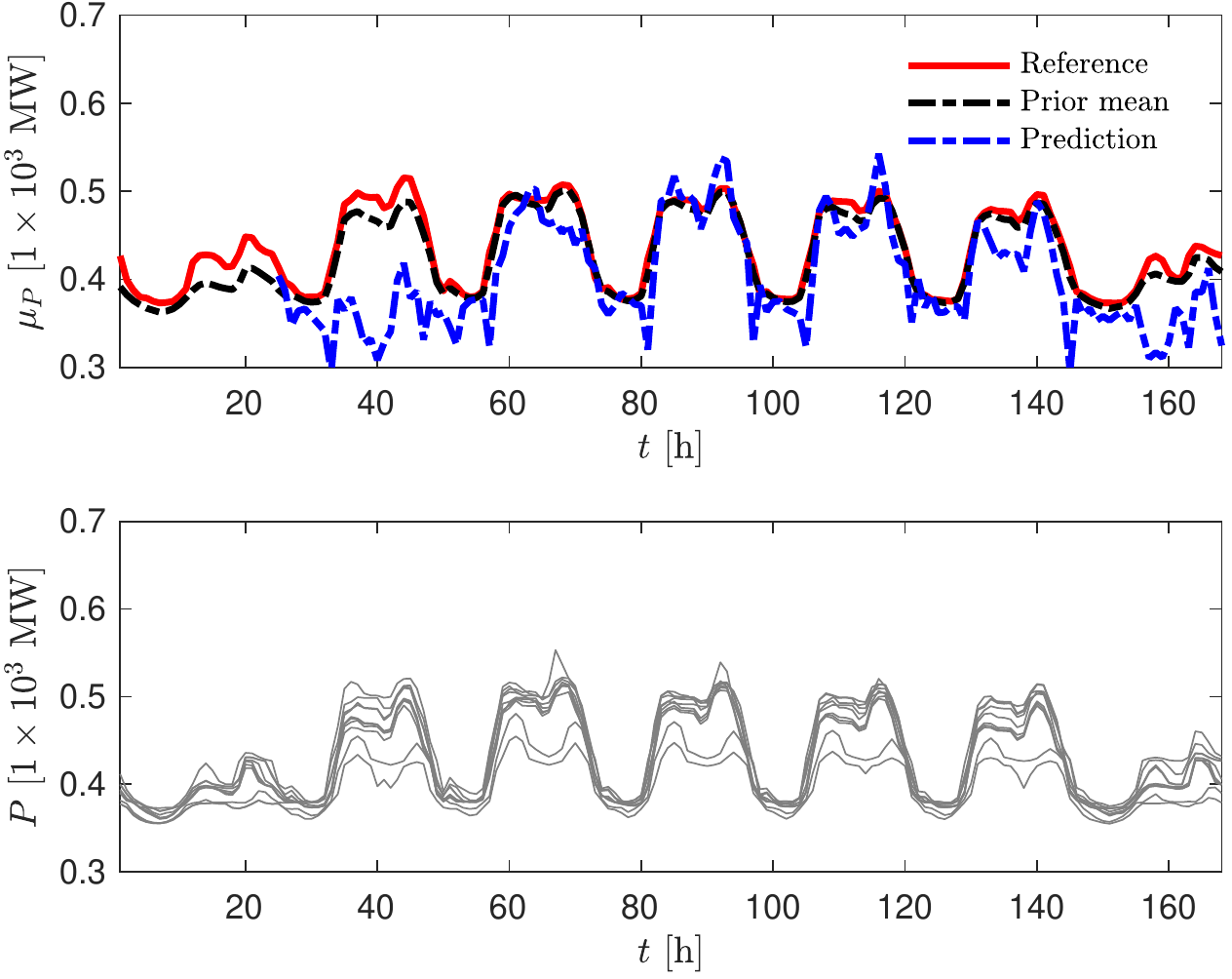}}%
  \caption{Weekly forecasts of the generation schedule for generator $15$ for four weeks using 24h observations on Monday.
    Top panel: Prediction (blue) compared against reference (red) and the prior mean (black).
    Bottom panel: Ensemble of $20$ timeseries used to compute the empirical covariance.}
  \label{fig:gen-forecast-wm} 
\end{figure*}

\subsection{Weekly forecasting of generation schedule}

Here, we use the EGPR method for weekly forecasting of the optimal power output of the Generator 15 for the weeks of 6/9--6/15, 8/11---8/17, 10/13--10/19, and 12/15--12/21 using the power output observation data from up to 20 previous weeks. 
Figures \ref{fig:gen-forecast-wom} and \ref{fig:gen-forecast-wm} show the comparison of the forecast and the reference results for five (Wednesday-Sunday) and six (Tuesday-Sunday) days, respectively.
As in the case with the total load forecast, we see that Monday data is weakly correlated with the rest of data. As a result, the optimal power output forecast is more accurate for Wednesday-Sunday using the previous Tuesday data than for the Tuesday-Sunday using the previous Monday data, in accordance with Figures~\ref{fig:load-forecast-wom} and \ref{fig:load-forecast-wm}.
An accurate Wednesday-Sunday forecast can be obtained using 10 previous weeks observations for computing the prior statistics, while 20 previous weeks  observations are required for an accurate Tuesday-Sunday forecast.
It can also be seen that Tuesday-Sunday forecasts for the weeks of 6/9--6/15 and 10/13--10/19 suffer from collapse to the prior mean, indicating poor forecasting performance.
The poorer performance of Tuesday-Sunday forecasts compared to Wednesday-Sunday forecasts highlights the importance of analyzing the historical data in order to properly construct the ensembles for EGPR.

\section{Discussion and conclusions}

The ensemble Gaussian process regression (EGPR) method is proposed and used for weekly forecasting of load demand and generation scheduling (optimal generator output) for a power grid with 700 buses and 134 generators using synthetic historical data.
The proposed EGPR method is based on the Gaussian process regression (GPR) method, with prior covariance matrices of the quantities of interest (QoI) computed from ensembles formed by up to twenty preceding weeks of QoI observations. 
To test the EGPR method, we generated hourly synthetic data for 365 days using historical Duke Energy hourly load profile to model load demand.
We use the EGPR method to compute weekly forecasts of the total load demand and scheduled real power generation for four weeks corresponding to the summer, fall, and early winter seasons.   
These numerical experiments demonstrate that in general the EGPR provides accurate forecasts and significantly outperforms the standard GPR and ARIMA forecasting methods. 
 
We found that Monday data for both load and scheduled real power generation is weakly correlated with data for the rest of the week.
As a result, the EGPR forecast for Wednesday-Sunday using the previous Tuesday data is more accurate than the forecast for Tuesday-Sunday using the previous Monday data.
An accurate Wednesday-Sunday forecast can be obtained using 10 previous weeks observation  for computing the prior statistics, while 20 previous weeks  observations are required for an accurate Tuesday-Sunday forecast.
The weak correlation between data for each Monday the data for the rest of each week is an artifact of the process employed for generating the synthetic data.

As shown by the numerical examples presented in Section~\ref{sec:results}, the choice of ensembles in critical to the performance of EGPR.
In future work we will explore the use of information criteria, which serve as proxies to the predictive accuracy of probabilistic forecasting methods~\cite{gelman-bayesian}, for selecting the ensemble that maximizes predictive accuracy.

\section{Acknowledgment}
This work was partially
supported by the U.S. Department of Energy (DOE) Office of
Science, Office of Advanced Scientific Computing Research
(ASCR) as part of the Multifaceted Mathematics for Rare,
Extreme Events in Complex Energy and Environment Systems
(MACSER) project. Pacific Northwest National Laboratory
 is operated by Battelle for the DOE
under Contract DE-AC05-76RL01830.

\bibliographystyle{ieeetr}
{}

\end{document}

%% file: hicss51-packages.tex
\usepackage[letterpaper]{geometry}
\usepackage{hicss51}
\usepackage{times}
\usepackage[none]{hyphenat}
\usepackage{url}
\usepackage{latexsym}
\usepackage{indentfirst}
\usepackage{graphicx}
\graphicspath{{images/}}